%% file: main.tex
\documentclass[10pt]{article}
\usepackage[utf8]{inputenc}
\usepackage{amsmath, amssymb, amsthm}
\usepackage{graphicx}
\usepackage{hyperref}
\usepackage{geometry}
\usepackage{times}  
\usepackage{natbib}  
\usepackage{authblk}  
\usepackage{anyfontsize} 

\usepackage{array}
\usepackage{tabu}
\usepackage{wrapfig}
\usepackage{float}

\usepackage{graphicx}
\usepackage{subcaption}
\usepackage{hyperref}
\usepackage{url}
\usepackage{xcolor}

\usepackage{multirow}

\newcommand{\mymbox}[1]{\mbox{\scriptsize #1}}
\newcommand{\quoteIt}[1]{``#1"}

\RequirePackage{algorithm}
\RequirePackage{algorithmic}

\newcommand{\TUremove}[1]{{}}
\input{math_commands.tex}

\author[1,3]{Xiaohui Tu \thanks{Corresponding author. Email addresses: 
\href{mailto:xiaohui.tu@hec.ca}{\textit{xiaohui.tu@hec.ca}},
\href{mailto:yossiri.adulyasak@hec.ca}{\textit{yossiri.adulyasak@hec.ca}},
\href{mailto:nima.akbarzadeh@hec.ca}{\textit{nima.akbarzadeh@hec.ca}}, and  \href{mailto:erick.delage@hec.ca}{\textit{erick.delage@hec.ca}}}}
\author[2]{Yossiri Adulyasak}
\author[1,3]{Nima Akbarzadeh}
\author[1,3]{Erick Delage}

\affil[1]{GERAD \& Department of Decision Sciences, HEC Montréal, Montr\'eal, H3T2A7, Canada}
\affil[2]{GERAD \& Department of Logistics and Operations Management, HEC Montréal, Montr\'eal, H3T2A7, Canada}
\affil[3]{MILA - Quebec AI Institute, Montréal, H2S3H1, Canada}

\title{Fair Resource Allocation in \\ Weakly Coupled Markov Decision Processes}

\date{}

\begin{document}
\maketitle

\begin{abstract} \normalsize
We consider fair resource allocation in sequential decision-making environments modeled as weakly coupled Markov decision processes, where resource constraints couple the action spaces of $N$ sub-Markov decision processes (sub-MDPs) that would otherwise operate independently. We adopt a fairness definition using the generalized Gini function instead of the traditional utilitarian (total-sum) objective. 
After introducing a general but computationally prohibitive solution scheme based on linear programming, we focus on the homogeneous case where all sub-MDPs are identical. For this case,  we show for the first time that the problem reduces to optimizing the utilitarian objective over the class of ``permutation invariant" policies. This result is particularly useful as we can exploit {efficient algorithms that optimizes the utilitarian objective such as Whittle index policies in restless bandits to solve the problem with this fairness objective. For more general settings, }
\TUremove{Whittle index policies in the restless bandits setting while, for the more general setting, }we introduce a count-proportion-based deep reinforcement learning approach. Finally, we validate our theoretical findings with comprehensive experiments, confirming the effectiveness of our proposed method in achieving fairness.
\end{abstract}

\section{INTRODUCTION}
Machine learning (ML) algorithms play a significant role in automated decision-making processes, influencing our daily lives. Mitigating biases within the ML pipeline is crucial to ensure fairness and generate reliable outcomes \citep{caton2020fairness}. Extensive research has been conducted to enhance fairness across various applications, such as providing job hiring services {\citep{van2020hiring, cimpean2024reinforcement}}, assigning credit scores and loans \citep{kozodoi2022fairness}, and delivering healthcare services \citep{farnadi2021individual, chen2023algorithmic}.  

However, most real-world decision processes are sequential in nature and past decisions may have a long-term impact on equity \citep{damour2020fairness}. For example, if people are unfairly denied credit or job opportunities early in their careers, there would be long-term consequences on opportunities for advancement \citep{liu2018delayed}. Another motivating example is taxi dispatching. If certain areas are consistently prioritized over others, then there can be long-term disparities in service accessibility. This may lead to long waiting times for passengers in certain neighborhoods, while taxis run empty and seek passengers in other areas {\citep{liu2021meta, guo2023fairness}}.

Fairness is a complex and multi-faceted concept, and there are many different ways in which it can be operationalized and measured. We resort to the generalized Gini social welfare function (GGF) \citep{weymark1981generalized}, which covers various fairness measures as special cases. The long-term impacts of fair decision dynamics have recently been approached using Markov decision processes (MDPs) \citep{wen2021algorithms, puranik2022dynamic, ghalme2022long}. {Studying fairness in MDPs helps mitigate bias and inequality in decision-making processes and evaluate their broader societal and operational impacts across diverse applications.}

To the best of our knowledge, we are the first to incorporate fairness considerations in the form of the GGF objective within weakly coupled Markov decision processes (WCMDPs) \citep{hawkins2003langrangian, adelman2008relaxations}, which can {be considered as an extension of restless multi-arm bandit problems (RMABs) \citep{hawkins2003langrangian, zhang2022near} to multi-action and multi-resource settings. This model is particularly relevant to resource allocation problems, as it captures the complex interactions of coupled MDPs (arms) over time restricted by limited resource availability,} and allows the applicability of our work to various applications in scheduling \citep{saure2012dynamic, el2024weakly}, application screening \citep{gast2024reoptimizationnearlysolvesweakly}, budget allocation \citep{boutilier2016budget}, and inventory \citep{el2024weakly}.

\textbf{Contributions \quad} Our contributions are as follows. \textit{Theoretically}, we reformulate the WCMDP problem with the GGF objective as a linear programming (LP) problem, and show that, under symmetry, it reduces to maximizing the average expected total discounted rewards, called the utilitarian approach.
\textit{Methodologically}, we propose a state count approach to further simplify the problem, and introduce a count proportion-based deep reinforcement learning (RL) method that can solve the reduced problem efficiently and can scale to larger cases by assigning resources proportionally to the number of stakeholders. \textit{Experimentally}, we design various experiments to show the GGF-optimality, flexibility, scalability and efficiency of the proposed deep RL approach. We benchmark our approach against the Whittle index policy on machine replacement applications modeled as RMABs \citep{akbarzadeh2019restless}, showing the effectiveness of our method in achieving fair outcomes under different settings.

There are two studies closely related to our work. 
The first work by \cite{gast2024reoptimizationnearlysolvesweakly} considers symmetry simplification and count aggregation MDPs. They focus on solving an LP model repeatedly with a total-sum objective to obtain asymptotic optimal solutions when the number of coupled MDPs is very large, whereas we explicitly address the fairness aspect and exploit a state count representation to design scalable deep RL approaches. The second work by \cite{siddique2020learning} integrates the fair Gini multi-objective RL to treat every user equitably. This fair optimization problem is later extended to the decentralized cooperative multi-agent RL by \cite{zimmer2021learning}, and further refined to incorporate preferential treatment with human feedback by \cite{siddique2023fairness} and {\cite{yu2023fair}}. In contrast, our work demonstrates that the WCMDP with the GGF objective and identical coupled MDPs reduces to a much simpler utilitarian problem, which allows us to exploit its structure to develop efficient and scalable algorithms. A more comprehensive literature review on fairness in {resource allocation,} MDPs, RL, and RMABs, is provided in Appendix \ref{appendix:literature} to clearly position our work.

\section{BACKGROUND}
We start by reviewing infinite-horizon WCMDPs and introducing the GGF for encoding fairness. We then define the fair optimization problem and provide an exact solution scheme based on linear programming.

\textbf{Notation \quad} Let $[N] := \{1, \dots, N\}$ for any integer $N$. For any vector $\vv\in\mathbb{R}^N$, the $n$-th element is denoted as $v_n$ and the average value as $\bar{v} = \frac{1}{N}\sum_{n=1}^N v_n$. An indicator function $\sI\{x\in A\}$ equals 1 if $x \in A$ and 0 otherwise. For any set $X$, $\Delta(X)$ represents the set of all probability distributions over $X$. We let $\sS^N$ be the set of all  $N!$ permutations of the indices in $[N]$ and $\G$ be the set of all permutation operators so that $Q\in\G$ if and only if there exists a $\sigma\in\sS^N$ such that $Q\vv(n)=\vv_{\sigma(n)}$ for all $n\in[N]$ when $\vv\in\mathbb{R}^N$.

\subsection{The Weakly Coupled MDP}

We consider $N$ MDPs indexed by $n \in \gN := [N]$ interacting in discrete-time over an infinite horizon $t \in \gT := \{0, 1, \dots, \infty\}$. The $n$-th MDP $\M_n$, also referred as sub-MDP, is defined by a tuple {$(\S_n, \A_n, \P_n, \R_n, \mu_n, \gamma)$}, where $\S_n$ is a finite set of states with cardinality $S$, and $\A_n$ is a finite set of actions with cardinality $A$. 
The transition probability function 
is defined as $\P_n(s'_n| s_n, a_n) = \Prob(s_{t+1, n} = s'_n|s_{t, n} = s_n, a_{t, n} = a_n)$, which represents the probability of reaching state $s'_n \in \S_{n}$ after performing action $a_n \in \A_{n}$ in state $s_n \in \S_{n}$ {at time $t$}.
The reward function $r_{n}(s_n, a_n)$ denotes the immediate real-valued reward obtained by executing action $a_n$ in state $s_n$. Although the transition probabilities and the reward function may vary with the sub-MDP $n$, we assume that they are stationary across all time steps for simplicity.
The initial state distribution is represented by $\mu_n\in\Delta(\S_n)$, and the discount factor, common to all sub-MDPs, is denoted by $\gamma \in [0, 1)$.

An infinite-horizon WCMDP $\Mn$ consists of $N$ sub-MDPs, where each sub-MDP is independent of the others in terms of state transitions and rewards. They are linked to each other solely through a set of $K$ constraints on their actions at each time step. Formally, the WCMDP is defined by a tuple $(\Sn, \An, \Pn, \Rn, \vmu, \gamma)$, where the state space $\Sn$ is the Cartesian product of individual state spaces, and the action space $\An$ is a subset of the Cartesian product of action spaces, defined as $\An := \{(a_1, \dots, a_N) \mid \sum_{n=1}^N d_{k, n}(a_n) \leq b_k, \forall k \in \gK, \ a_n \in \A_{n}\},$ where $\gK := [K]$ is the index set of constraints, $d_{k, n}(a_n) \in \sR_{+}$ represents the consumption of the $k$-th resource consumption by the $n$-th MDP when action $a_n$ is taken, and $b_k\in\mathbb{R}_+$ the available resource of type $k$.\footnote{Actually, $b_k\leq \sum_{n=1}^N\max_{a_n\in\A_n}d_{k,n}(a_n)$, w.l.o.g.} We define an idle action that consumes no resources for any resource $k$ to ensure that the feasible action space is non-empty.

The state transitions of the sub-MDPs are independent, so the system transits from state $\s$ to state $\s'$ for a given feasible action $\a$ at time $t$ with probability $\Pn (\s'|\s, \a) = \prod^{N}_{n=1} p_n(s'_{n} | s_{ n}, a_{n}) = \prod^{N}_{n=1} \Prob(s_{t+1, n} = s'_n|s_{t, n} = s_n, a_{t, n} = a_n)$. After choosing an action $\a \in \An$ in state $\s \in \Sn$, the decision maker receives rewards defined as $\Rn(\s, \a) = (r_{1}(s_1, a_1), \dots, r_{N}(s_N, a_N))$ with each component representing the reward associated with the respective sub-MDP $\M_n$. We employ a vector form for the rewards to offer the flexibility for formulating fairness objectives on individual expected total discounted rewards in later sections.

We consider stationary Markovian policy $\vpi: \Sn \times \An \rightarrow [0,1]$, with notation $\vpi(\s, \a)$ capturing the probability of performing action $\a$ in state $\s$. The initial state $\s_0$ is sampled from the distribution $\vmu$. 
Using the discounted-reward criteria, the state-value function $V^{\vpi}_n$ specific to the $n$-th sub-MDP $\mathcal{M}_n$, starting from an arbitrary initial state $\s_0$ under policy $\vpi$, is defined as $V^{\vpi}_{n} (\s_0):= \E_{\vpi}\left[ \sum_{t=0}^{\infty} \gamma^t r_{n}(s_{t,n}, a_{t,n}) |\s_0 \right],$ where $\a_t \sim \vpi(\s_t, \cdot)$. The joint state-value vector-valued function $\V^{\vpi}(\s_0): \Sn \rightarrow \sR^N$ is defined as the column vector of expected total discounted rewards for all sub-MDPs under policy $\vpi$, i.e., $\V^{\vpi}(\s_0) := \left( V^{\vpi}_{1}(\s_0), V^{\vpi}_{2}(\s_0), \ldots, V^{\vpi}_{N}(\s_0) \right)^\top$. We define {$\Vmu$} as the expected vectorial state-value under initial distribution $\vmu$, i.e.,
\setlength{\abovedisplayskip}{3pt}
\setlength{\belowdisplayskip}{3pt}
\begin{equation}
\Vmu := \E[\V^\vpi(\s_0)|\s_0\sim{\vmu}].
\label{eq:V}
\end{equation}

\subsection{The Generalized Gini Function}\label{sec:GGF}

{The vector $\Vm$ represents the {expected} utilities for {sub-MDPs}.} A social welfare function aggregates these utilities into a scalar, measuring fairness in utility distribution with respect to a maximization objective.

Social welfare functions can vary depending on the values of a society, such as $\alpha$-fairness \citep{mo2000fair}{, Nash social welfare \citep{fan2022welfare, mandal2022socially}}, or max-min fairness {\citep{bistritz2020my, cousins2022fair}}. Following \cite{siddique2020learning}, we require a fair solution to meet three properties: efficiency, impartiality, and equity. \TUremove{\textit{Efficiency} implies that no user's welfare can improve without impacting the others. \textit{Impartiality} implies that identical {sub-MDPs} are treated similarly regardless of utility vector permutations. \textit{Equity} favors more balanced solutions, reducing differences among {sub-MDPs}.} 
{This {motivates} the use of GGF from economics \citep{weymark1981generalized},} which satisfies these properties. For $N$ {sub-MDPs}, GGF is defined as
\(\GGF_{\vw}[\vv]:=\min_{\sigma\in\sS^N}\sum_{n=1}^N {w}_{n} {v}_{\sigma(n)}\), where $\vv \in \sR^N$, $\vw \in \Delta(\gN)$ is non-increasing in $n$, i.e., $w_1 \geq w_2 \geq \cdots \geq w_N$.
{Intuitively, since $\GGF_{\vw}[\vv]=\sum_{n=1}^N {w}_n {v}_{\sigma^*(n)}$ with $\sigma^*$ as the minimizer, which reorders the terms of $\vv$ from lowest to largest, it computes the weighted sum of $\vv$ assigning larger weights to its lowest components.} When the order of sub-MDPs is fixed, we use the equivalent formulation $\GGF_{\vw}[\vv]=\min_{\sigma\in\mathbb{S}^N}\sum_n w_{\sigma(n)} v_n$ as permuting either vector results in the same outcome.

As discussed in \cite{siddique2020learning}, {GGF can reduce to special cases by setting its weights to specific values,}
including the maxmin egalitarian approach ($w_{1} \rightarrow 1, w_{2} \rightarrow 0, \ldots, w_{N} \rightarrow 0$) \citep{rawls1971atheory}, regularized maxmin egalitarian ($w_{1} \rightarrow 1, w_{2} \rightarrow \epsilon, \ldots, w_{N} \rightarrow \epsilon$), leximin notion of fairness ($w_{k} /w_{k+1} \rightarrow \infty$) \citep{rawls1971atheory, moulin1991axioms}, and the utilitarian approach formally defined below for the later use in reducing the GGF problem.
\begin{definition}[Utilitarian Approach]\label{def:utilitarian}
The utilitarian approach within the GGF framework is obtained by setting equal weights for all individuals, i.e., $\vw_{\voneN} := \vone/N$ so that $\GGF_{\vw_{\voneN}}[\vv]=\min_{\sigma\in \sS^N}\sum_{n=1}^N \frac{1}{N} {v}_{\sigma(n)} = \frac{1}{N}\sum_{n=1}^N v_n = \bar{v}.$
\end{definition}

The utilitarian approach maximizes average utilities over all individuals but does not guarantee fairness in utility distribution, as some {sub-MDPs} may be disadvantaged to increase overall utility. The use of GGF offers flexibility by encoding various fairness criteria in a structured way. Moreover, $\GGF_{\vw}[\vv]$ is concave in $\vv$, which has nice properties for problem reformulation.

\subsection{The GGF-WCMDP Problem}
{By combining GGF and the vectored values from  the WCMDP in (\ref{eq:V}), the goal of the GGF-WCMDP problem (\ref{eq:ggf}) is defined as finding a stationary policy $\vpi$ that maximizes the GGF of the expected total discounted rewards, i.e., $\underset{\vpi}{\max}  \GGF_{\vw}\left[\Vm\right]$ that is equivalent to
\begin{equation}
\underset{\pi}{\max} \min_{\sigma \in \sS_N} {\vw}^\top_\sigma
\E_{\vpi}\left[ \sum_{t=0}^{\infty} \gamma^t \r(\s_t, \a_t) \bigg| \s_0 \sim \vmu \right].
\label{eq:ggf}
\end{equation}}{We note that Lemma 3.1 in \cite{siddique2020learning} establishes the optimality of stationary Markov policies for any multi-objective discounted infinite-horizon MDP under the GGF criterion. {To obtain an }optimal policy for the GGF-WCMDP problem (\ref{eq:ggf}){, we introduce}\TUremove{ can be computed by solving} the following LP model with the GGF objective (GGF-LP):} 

\begingroup\allowdisplaybreaks
\begin{subequations}\label{ggf-mdp-d}

{\begin{eqnarray}
\max_{\bm{\lambda},\bm{\nu},\bm{q}} && \hskip-1em\sum_{i=1}^N \lambda_i + \sum_{j=1}^N \nu_j \TUremove{\qquad \mbox{(GGF-LP)}}\\
 \mathrm{s.t.} &&  \hskip-1em \lambda_i + \nu_j \leq w_i \sum \limits_{\s \in \Sn} \sum \limits_{\a \in \An} \r_j(\s, \a) q(\s, \a), \, \forall i,j {\in \gN}, \\
 &&  \hskip-2em \sum\limits_{\a \in \An} q(\s, \a) -  \gamma \sum \limits_{\s^\prime \in \Sn} \sum \limits_{\a \in \An} 
 q(\s^\prime, \a) \Pn(\s | \s^\prime, \a) = \vmu(\s), \quad \forall \s \in \Sn, \label{eq:ggf-mdp-d:c2} \\
 &&  \hskip-2em q(\s, \a) \geq 0, \quad \relax \forall \s \in \Sn, \quad \forall \a \in \An.\label{eq:ggf-mdp-d:c3}
\end{eqnarray}}
\end{subequations}
\endgroup
\vspace{3pt}

See Appendix \ref{A:dual-lp} \TUremove{and \citep{doi:10.1287/moor.1110.0516}} for details on obtaining model (\ref{ggf-mdp-d}) that exploits the dual linear programming formulation for solving discounted MDPs. Here, $q(\s, \a)$ represents the total discounted visitation frequency for state-action pair $(\s, \a)$, starting from $\s_0$.

{The dual form separates dynamics from rewards, with the expected discounted reward for sub-MDP $n$ given by $\sum_{\s \in \Sn} \sum_{\a \in \An} \r_n(\s, \a) q(\s, \a)$. The one-to-one mapping between the solution $q(\s, \a)$ and an optimal policy $\vpi(\s, \a)$ is
\(
  \vpi(\s, \a) = q(\s,\a)/\sum_{\a \in \An} q(\s,\a)
\).} Scalability is a critical challenge in obtaining exact solutions {as the state and action spaces grow exponentially with respect to the number of sub-MDPs{, making the problem intractable}.} We thus explore approaches that exploit symmetric problem structures, apply count-based state aggregation, and use RL-based approximation methods, to address this scalability issue, which will be discussed next.

\section{UTILITARIAN REDUCTION UNDER SYMMETRIC SUB-MDPS}\label{sec:reduction}

{In Section \ref{sec:symmetric-WCMDP}, we will formally define the concept of symmetric WCMDPs (definition \ref{def:s-WCMDP}) and prove that an optimal policy of the GGF-WCMDP problem can be obtained by solving the utilitarian WCMDP using \quoteIt{permutation invariant} policies. This enables the use of Whittle index policies in the RMAB setting while, for the more general setting, Section \ref{sec:count-MDP} proposes a count aggregation MDP reformulation that will be solved using deep RL in Section \ref{sec:count-based-NN}.

\subsection{GGF-WCMDP Problem Reduction} \label{sec:symmetric-WCMDP}
We start with formally defining the conditions for a WCMDP to be considered symmetric.

\begin{definition}[Symmetric WCMDP]\label{def:s-WCMDP}
\par A WCMDP is symmetric if
\vspace{-6pt}
\begin{enumerate}[leftmargin=*]
\setlength{\parskip}{3pt}
\item \textbf{(Identical Sub-MDPs)} Each sub-MDP is identical, i.e.,  $\S_n = \S$, $\A_{n} = \A$, $\P_{n} = \P$, $\R_{n} = \R$, $\mu_{n} = \mu$, for all $n\in \gN$, and for some $(\S, \A, \P, \R, \mu, \gamma)$ tuple.
\item \textbf{(Symmetric resource consumption)} For any $k\in\mathcal{K}$, the number of resources consumed is the same for each sub-MDP, i.e., {$d_{k,n}(a_n)=d_{k}(a_n)$} for all $n\in \gN$, and for some $d_{k}(\cdot)$.
\item \textbf{(Permutation-Invariant Initial Distribution)}  For any permutation operator $Q \in \G$, the probability of selecting the permuted initial state $Q \so$ is equal to that of selecting $\so$, i.e., $\vmu(\so) = \vmu(Q \so), \forall \so \in \Sn, \forall Q \in \G$.
\end{enumerate}
\end{definition}

The conditions of symmetric WCMDP identify a class of WCMDPs that is invariant under any choice of indexing for the sub-MDPs. This gives rise to the notion of \quoteIt{permutation invariant} policies (see definition 1 in \cite{9683491}) and the question of whether this class of policies is optimal for symmetric WCMDPs.

\begin{definition}[Permutation Invariant Policy]\label{def:permutation-invariant-policy}
A Markov stationary policy $\vpi$ is said to be permutation invariant if the probability of selecting action $\a$ in state $\s$ is equal to that of selecting the permuted action $Q\a$ in the permuted state $Q\s$, for all $Q \in \G$. Formally, this can be expressed as $\vpi(\s, \a) = \vpi(Q\s, Q \a)$, for all $Q \in \G$, $\s \in \Sn$ and $\a \in \An$. 
\end{definition}

{This symmetry ensures that the expected state-value function, when averaged over all trajectories, is identical for each sub-MDP, leading to a uniform state-value representation. From this observation, and applying Theorem 6.9.1 from \citep{puterman2014markov}, we construct a permutation-invariant policy from any policy, resulting in uniform state-value representation (Lemma \ref{thm:bar-policy}).}

\begin{lemma}[Uniform State-Value Representation]\label{thm:bar-policy}
If a WCMDP is symmetric, then for any policy $\vpi$, there exists a corresponding permutation invariant policy $\bpi$ such that the vector of expected total discounted rewards for all sub-MDPs under $\bpi$ is equal to the average of the expected total discounted rewards for each sub-MDP, i.e., $\V^{\bpi}_0
= \frac{1}{N} \sum_{n=1}^N V_{0,n}^\vpi
\vone.$
\end{lemma}

{The proof is detailed in Appendix \ref{A:bar-policy}. Furthermore, one can use the above lemma to show that the optimal policy for the \TUremove{optimization}{GGF-WCMDP problem (\ref{eq:ggf}) under symmetry }can be recovered from solving the \TUremove{optimization}problem with equal weights, i.e., the utilitarian approach.} Our main result is presented in the following theorem.
See Appendix \ref{A:eqv} for a detailed proof.

\begin{theorem}[Utilitarian Reduction] \label{thm:eqv}
{For a symmetric WCMDP, let $\Pi^*_{{\voneN,\mymbox{PI}}}$ be the set of optimal policies for the utilitarian approach that is permutation invariant, then $\Pi^*_{\voneN,\mymbox{PI}}$ is necessarily non-empty and all $\pi^*_{\voneN,\mymbox{PI}}\in\Pi^*_{\voneN,\mymbox{PI}}$ satisfies 
\[
\GGF_{\vw}[\V_0^{\pi_{\voneN,\mbox{\tiny{PI}}}^{*}}]
= \max_{\vpi} \GGF_{\vw}\left[\Vm \right], \forall \vw \in\Delta(N).
\]}
\end{theorem}
{This theorem simplifies solving the GGF-WCMDP problem by reducing it to an equivalent utilitarian problem, showing that at least one permutation-invariant policy is optimal for the original GGF-WCMDP problem and the utilitarian reduction. Therefore, we can restrict the search for optimal policies to this specific class of permutation-invariant policies. 
The utilitarian approach does not compromise the GGF optimality and allows us to leverage more efficient and scalable techniques to solve the GGF-WCMDP problem (Eq. \ref{eq:ggf}), such as the Whittle index policies for RMABs, as demonstrated in the experimental section. {We note that the utilitarian reduction theorem can be extended to a broader class of fairness measures, such as $\alpha$-fairness, as long as the fairness measure is concave, permutation invariant, and constant vector invariant (Corollary \ref{cr:utilitarian}). See Appendix \ref{appendix:extension} for a detailed proof.}

\subsection{The Count Aggregation MDP}\label{sec:count-MDP}
{Assuming symmetry across all $N$ sub-MDPs and using a permutation-invariant policy within a utilitarian framework allows us to simplify the global MDP by aggregating the sub-MDPs based on their state counts and tracking the number of actions taken in each state. Since each sub-MDP follows the same transition probabilities and reward structure, we can represent the entire system more compactly.} {This symmetry consideration is practical in many real-world applications where a large number of identical or interchangeable identities demand fair and efficient treatment, such as patients in healthcare or taxi drivers in public transportation services. By leveraging symmetry, we can reduce computational complexity for scalable fair solutions while inherently enforcing fairness as the policy treats all sub-MDPs equivalently.}

Motivated by the symmetry simplification representation in \cite{gast2024reoptimizationnearlysolvesweakly} for the utilitarian objective, we consider an aggregation $\phi=(f, g_s)$, where $f: \Sn \rightarrow \sN^{S}$ maps state $\s$ to a count representation $\x$ with $x_s$ denoting the number of sub-MDPs in the $s$-th state. Similarly, $g_s: \An \rightarrow \sN^{S \times A}$ maps action $\a$ to a count representation $\u$, where $u_{s, a}$ indicates the number of MDPs at $s$-th state that performs $a$-th action. We can formulate the count aggregation MDP (definition \ref{def:count-MDP}). The details on obtaining the exact form are in Appendix \ref{def:countM}.

\begin{definition}[Count Aggregation MDP]\label{def:count-MDP} The count aggregation MDP $\cM$ derived from a WCMDP $(\Sn, \An,$\\$ \Pn, \Rn, \vmu, \gamma)$ consists of the elements $(\cS, \cA, \cP, \bR, \cmu, \gamma)$. \label{def:countMDP}
\end{definition}

Both representations lead to the same optimization problem as established in \cite{gast2024reoptimizationnearlysolvesweakly} when the objective is utilitarian. Using the count representation, the mean expected total discounted reward $\bV^{{\vpi}_{\voneN}}$ for a WCMDP $\Mn$ with permutation invariant distribution $\vmu$ 
and equal weights $\vw_{\voneN}$ (Theorem \ref{thm:eqv})
is then equivalent to the expected total discounted mean reward $\bV^{\cpi}$ for the count aggregation MDP $\cM$ given the policy $\cpi: \cS \rightarrow \Delta(\cA)$ under aggregation mapping with initial distribution $\cmu$, i.e., $\bV^{{\vpi}_{\voneN}}
    = \frac{1}{N}\sum_{n=1}^N V^{{\vpi}_{\voneN}}_{0,n} = \frac{1}{S} \sum_{s=1}^S V^{{\vpi}_{\phi}}_{0,s} 
    = \bV^{\cpi}$.

The objective in Eq. \ref{eq:ggf} is therefore reformulated as $\underset{\cpi}{\max} \, \bV^{\cpi}$, i.e.,
\begin{equation}
\underset{\cpi}{\max} \, \frac{1}{S} 
\E_{\cpi}\left[ \sum_{t=0}^{\infty} \gamma^t \bar{r}_{\phi}(\x_t, \u_t) \bigg| \x_0 \sim \cmu \right].
\label{eq:ggf-eq}
\end{equation}

An LP method is provided to solve the count aggregation MDP in Appendix \ref{A:count-dual-lp}.

\section{COUNT-PROPORTION-BASED DRL}\label{sec:count-based-NN}
We now consider the situation where the transition dynamics $\Pn_\phi$ are unknown and the learner computes the (sub-)optimal policy through trial-and-error interactions with the environment.
{In Section \ref{sec:policy-nn}, we introduce a count-proportion-based deep RL (CP-DRL) approach. This method incorporates a stochastic policy neural network with fixed-sized inputs and outputs, designed for optimizing resource allocation among stakeholders under constraints with count representation. In Section \ref{sec:sampling}, we detail the priority-based sampling procedure used to generate count actions.}

\subsection{Stochastic Policy Neural Network} \label{sec:policy-nn}
One key property of the count aggregation MDP is that the dimensions of the state space $\cS$ and the action space $\cA$ are constant and irrespective of the number of sub-MDPs. To further simplify the analysis and eliminate the influence of $N$, we define the count state proportion as $\dx = \x/N$ and the resource proportion constraint for each resource $k$ as 
$\bar{b}_k= b_k/(N\max_{a\in\A}d_k(a))\in[0,1]$. \TUremove{\EDcomment{Here, you need to define: $\bar{b}_k= b_k/(N\max_{a\in\A}d_k(a))\in[0,1]$.} Thanks for putting this, integrated.}

{This converts the states into a probability distribution, allowing generalization when dealing with a large number of agents. The stochastic policy network in Figure \ref{fig:stochastic-policy-network} is designed to handle the reduced count aggregation MDP problem (\ref{eq:ggf-eq}) by transforming the tuple $(\dx, \dvb)$ into a priority score matrix $\mU$ and a resource-to-use proportion vector $\pb$, which are then used to generate count actions $\u$ via a sampling procedure (discussed in Section \ref{sec:sampling}).}

\begin{figure}[htb]
    \centering
    \includegraphics[width=0.6\textwidth]{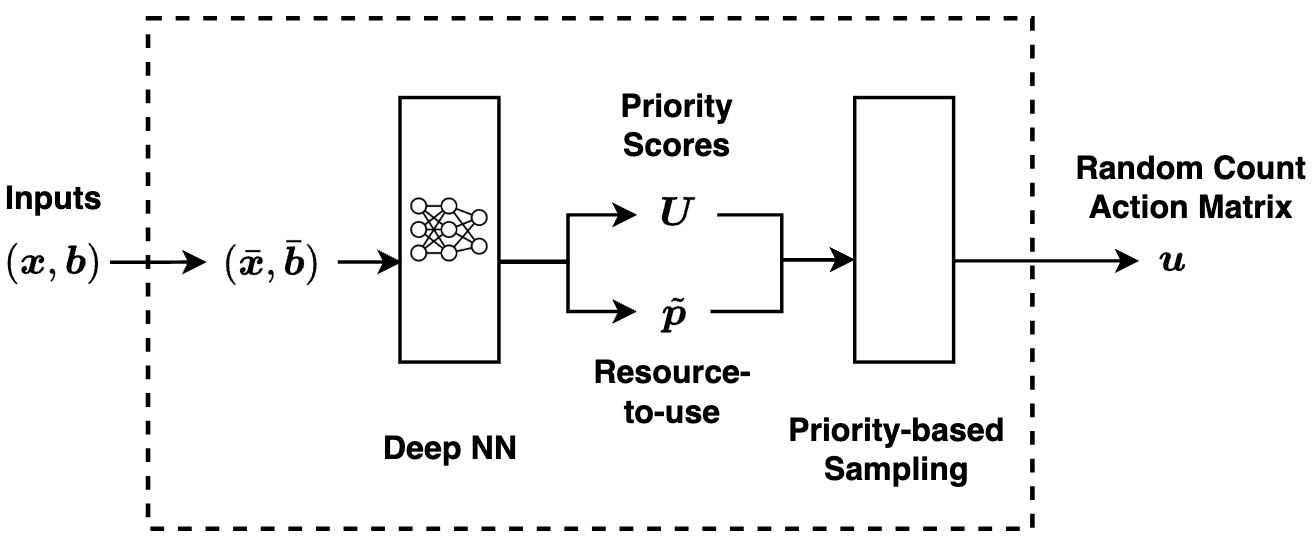}
    \caption{CP-based Stochastic Policy Neural Network\TUremove{\EDcomment{In figure, replace "NN Blocks" with "Deep NN", and "Sampling" with "Priority-based sampling", also "Output" should be "Random count action matrix"} Good point, replaced}}
    \label{fig:stochastic-policy-network}
\end{figure}

{The policy network features fixed-size inputs and outputs, enabling scalability in large-scale systems without requiring structural modifications when adjusting the number of resources or machines. The input consists of a fixed-size vector of size $S+K$, combining the count state proportion $\dx \in [0, 1]^S$ and the resource proportion $\dvb \in [0, 1]^K$. The policy network processes these inputs to produce outputs of size $S \times A + K$, which include a matrix $\mU \in (0, 1]^{S \times A}$ representing the priority scores for selecting count actions and a vector $\pb \in [0, 1]^K$ representing the proportion of resource usage relative to the total available resources $\dvb$.} 

The advantages of adding additional resource proportion nodes $\pb$ to the output layer are twofold.
First, it reduces the computational effort required to ensure that the resource-to-use does not exceed $\dvb$. Instead, the resource constraint is satisfied by restricting the resource-to-use proportions $\pb$ to be element-wise proportional to $\dvb$. Second, since the optimal policy may not always use all available resources, we incorporate the additional nodes to capture the complex relationships between different states for more effective strategies to allocate resources.

\subsection{Priority-based Sampling Procedure} \label{sec:sampling}

{Priority-based sampling presents a challenge since legal actions depend on state and resource constraints. To address this, a masking mechanism prevents the selection of invalid actions. Each element $u^p_{s, a} \in \mU$ represents the priority score of taking {the $a$-th action in the $s$-th state}\TUremove{$a$ in state $s$}.} When the state count $x_s$ is zero, it implies the absence of sub-MDPs in this state, and the corresponding priority score is masked to zero. Legal priorities are thus defined for states with non-zero counts, i.e., $u^p_{s,a} = 0$ if $x_s = 0$ for all $s \in [S], a \in [A]$.

Since the selected state-action pairs must also satisfy multi-resource constraints, we introduced a forbidden set $\gF$, which specifies the state-action pairs that are excluded from the sampling process. The complete procedure is outlined in Algorithm \ref{alg:sampling}. The advantage of this approach is that the number of steps does not grow exponentially with the number of sub-MDPs. In the experiments, after obtaining the count action $\u$, a model simulator is used to generate rewards and the next state as described in Algorithm \ref{alg:simulation} in Appendix~\ref{apdx:algorithm}. The simulated outcomes are used for executing policy gradient updates and estimating state values.

\begin{algorithm}[htb]
\caption{Count Action Sampling Based on Priority Scores}
\label{alg:sampling}
\begin{algorithmic}
\STATE {\bfseries Input:} Count state $\x$, priority score matrix $\mU$, resource limitations $\vb$, resource-to-use proportion $\pb$, resource consumption function $\vd(a)$
\STATE \textbf{Initialize}: $\tilde{\vb} \leftarrow \vb \cdot \pb$, $\u \leftarrow \bm{0}_{S \times A}$, $\gF \leftarrow \emptyset$
\STATE Apply masking to $\mU$ and update the forbidden set by
$\gF \leftarrow \mathcal{F} \cup \{(s, a) \mid u^p_{s,a} = 0 \text{ for } s \in [S] \text{ and } a \in [A]\}$
\WHILE{$|\gF| < S\times A$}
\STATE Sample a state-action index pair $(s, a) \notin \gF$ with the probability proportional to $\mU$ 
    \IF{$d_{k}(a) \leq \tilde{b}_k$ for all $k$}
    \STATE Update $u_{s, a} \leftarrow u_{s, a} + 1$, $x_s \leftarrow x_s - 1$
    \STATE Update $\tilde{b}_k \leftarrow \tilde{b}_k - d_k(a)$ for all $k$
        \IF{$x_s =0$}
        \STATE Add all actions for the $s$-th state to forbidden set: $\gF \leftarrow \mathcal{F} \cup \{(s, a) \mid \forall a \in [A]\}$
        \ENDIF
    \ELSE 
    \STATE Add $(s,a)$ to forbidden set $\gF \leftarrow \mathcal{F} \cup \{(s, a)\}$
    \ENDIF
\ENDWHILE
\STATE {\bf Return:} Count action matrix $\u$
\end{algorithmic}
\end{algorithm}

One critical advantage of using CP-DRL is its \textit{scalability}. More specifically, the approach is designed to handle variable sizes of stakeholders $N$ and resources $K$ while preserving the number of aggregated count states constant for a given WCMDP. By normalizing inputs to fixed-size proportions, the network can seamlessly adapt to different scales, {making it highly adaptable.}
Moreover, the fixed-size inputs allow \textit{flexibility} that the neural network is trained once and used in multiple tasks with various numbers of stakeholders and resource limitations.

\section{EXPERIMENTAL RESULTS}

We apply our methods to the machine replacement problem \citep{delage2010percentile, akbarzadeh2019restless}, providing a scalable framework for evaluating the CP-DRL approach as problem size and complexity increase. We focus on a single resource ($K = 1$) and binary action ($A = 2$) for each machine, allowing validation against the Whittle index policy for RMABs \citep{whittle1988restless}. {We applied various DRL algorithms, including \textit{Soft Actor Critic} (SAC), \textit{Twin Delayed DDPG} (TD3), and \textit{Proximal Policy Optimization} (PPO). Among these, PPO algorithm \citep{schulman2017proximal} consistently delivers the most stable and high-quality performance. We thus choose PPO as the main algorithm for our CP-DRL approach (see Section \ref{sec:count-based-NN}). Our code is provided on GitHub.\footnote{
\url{https://github.com/x-tu/GGF-wcMDP}.}}

\textbf{Machine Replacement Problem \quad} The problem consists of $N$ identical machines following Markovian deterioration rules with $S$ states representing aging stages. The state space $\Sn$ is a Cartesian product. At each decision stage, actions $\a$ are applied to all machines under resource constraints, with action $a_n$ representing operation (\textit{passive} action) or replacement (\textit{active} action).
Resource consumption $d_{n}(a_n)$ is 1 for replacements and 0 for operations, with up to $b$ replacements per time step. The costs range from 0 to 1, transformed to fit the {reward representation} by multiplying by -1 and adding 1. Machines degrade if not replaced and remain in state $S$ until replaced. Refer to Appendix \ref{appendix:mrp-parameter} for cost structures and transition probabilities. We choose operational and replacement costs across two presets to capture different scenarios (see Appendix \ref{appendix:cost} for details): \textit{i) Exponential-RCCC} and \textit{{ii)} Quadratic-RCCC}.
 
The goal is to find a fair policy that maximizes the GGF score over the expected total discounted mean rewards with count aggregation MDP. {In cases like electricity or telecommunication networks \citep{nadarajah2024self}, where equipment is regionally distributed, a fair policy guarantees equitable operations and replacements, thereby preventing frequent failures in specific areas that lead to unsatisfactory and unfair results for certain customers}.

\textbf{Experimental Setup \quad} We designed a series of experiments to test the GGF-optimality, flexibility, scalability, and efficiency of our CP-DRL algorithm. We compare against {seven benchmarks, including} optimal solutions (OPT) from the GGF-LP model (\ref{ggf-mdp-d}) for small instances solved with Gurobi 10.0.3, the Whittle index policy (WIP) for RMABs, and a random (RDM) agent that selects actions randomly at each time step and averages the results over 10 independent runs. {Additionally,  we implemented a simple DRL baseline, Vanilla-DRL (V-DRL), with a utilitarian objective. The stochastic policy network employs a fully connected neural network that maps the vector $\s$ to a $N$-dimensional probability vector. We also implemented two heuristics to complement the random agent approach. The oldest-first (OFT) approach selects the machine in the worst state, while the myopic (MYP) selects the machine that maximizes immediate reward. We finally implemented an equal-resources (EQR) approach based on \cite{li2022efficient}, which imposes that each machine be replaced once every $N$ steps to ensure an equal distribution of resources.}

GGF weights decay exponentially with a factor of 2, defined as $\vw_n = 1/2^n$, and normalized to sum to 1. We use a uniform distribution $\vmu$ over $\Sn$ and set the discount factor $\gamma = 0.95$. We use Monte Carlo simulations to evaluate policies over $M$ trajectories truncated at time length $T$. We choose $M$ = 1,000 and $T$ = 300 across all experiments. Hyperparameters for the CP-DRL algorithm are in Appendix \ref{appendix:hyperparameter}.

\textbf{Experiment 1 (GGF-Optimality) \quad} We obtain optimal solutions using the OPT model for instances where $N \in \{3, 4, 5\}$, with each machine having $S=3$ states.
We select indexable instances to apply the WIP method for comparison. Note that, the WIP method is particularly effective in this case as it solves the equivalent utilitarian problem (as demonstrated in the utilitarian reduction result in Section \ref{sec:symmetric-WCMDP}). In most scenarios with small instances, WIP performs near-GGF-optimal since resources are assigned impartially, making it challenging for CP-DRL to consistently outperform WIP. As shown in Figure \ref{fig:optimality}, the CP-DRL algorithm converges toward or slightly below the OPT values across the scenarios for the \textit{Exponential-RCCC} case. WIP performs better than the random agent but does not reach the OPT values, especially as the number of machines increases. {CP-DRL either outperforms or has an equivalent performance as WIP but consistantly outperforms the random policy.}

\begin{figure*}[htb] 
    \centering
    \begin{minipage}{0.32\textwidth}
        \centering
        \includegraphics[width=\linewidth]{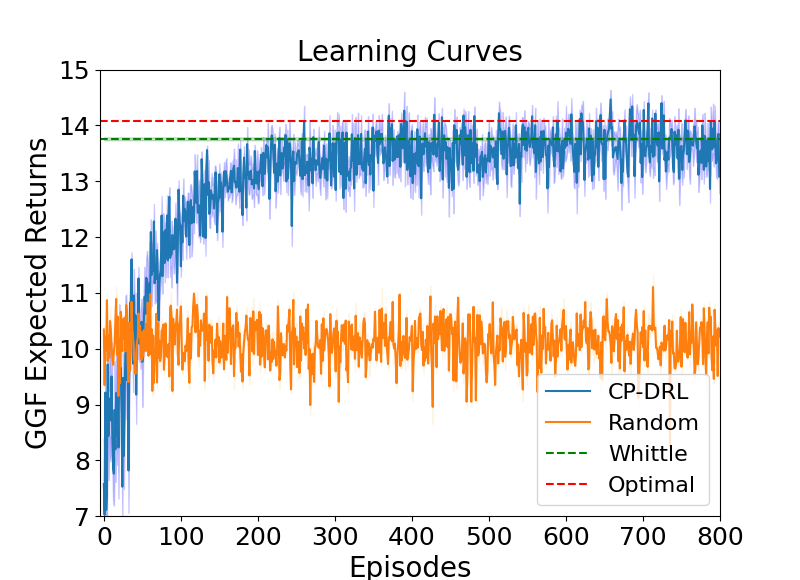}
        \caption*{(a) $N$=3}
    \end{minipage}
    \begin{minipage}{0.32\textwidth}
        \centering
        \includegraphics[width=\linewidth]{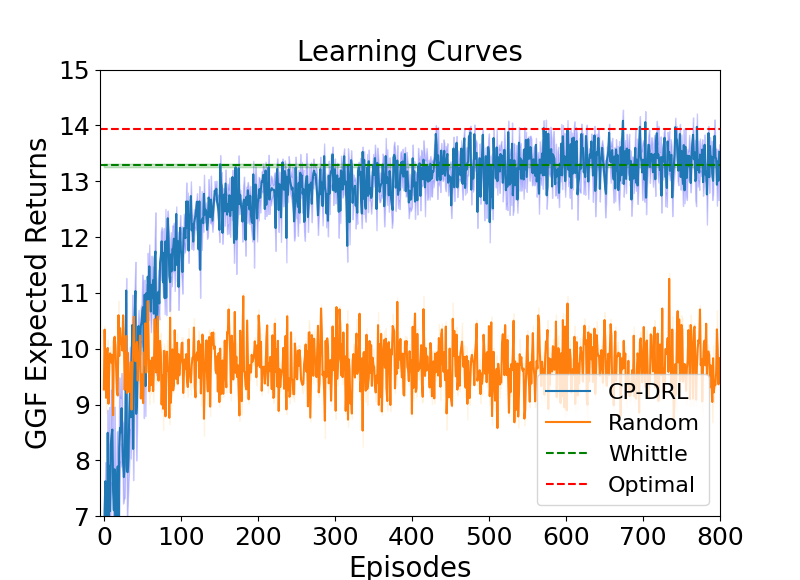}
        \caption*{(b) $N$=4}
    \end{minipage}
    \begin{minipage}{0.32\textwidth}
        \centering
        \includegraphics[width=\linewidth]{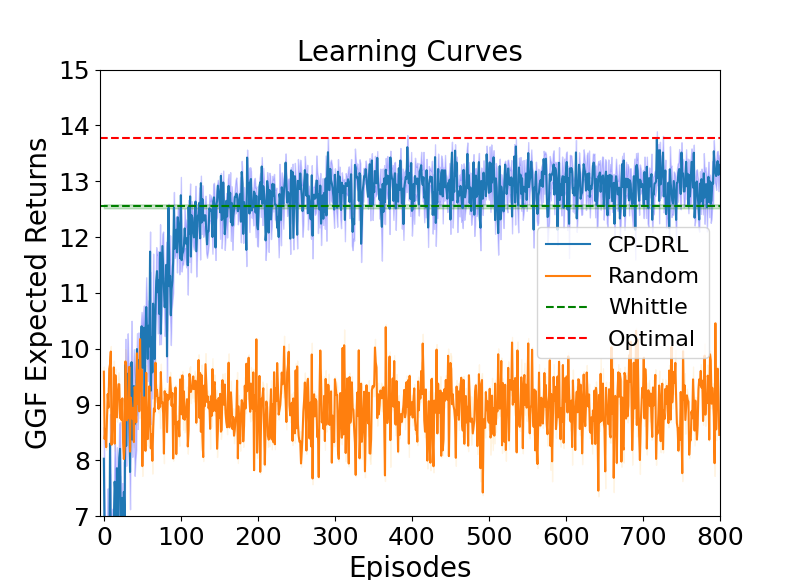}
        \caption*{(c) $N$=5}
    \end{minipage}
    \caption{{\textbf{(Colored) Learning Curves for Different Numbers of Machines ($N$ from 3 to 5).} \quad Experimental results for the \textit{Exponential-RCCC} scenario are shown with y-axes starting at 7 for zoom-in. Red dashed lines represent the OPT values, green dashed lines show the WIP performance, blue lines depict CP-DRL learning curves over 800 episodes, and orange lines show the RDM performance. Shaded areas indicate the standard deviation across 5 runs.}}
    \label{fig:optimality}
\end{figure*}

\textbf{Experiment 2 (Flexibility) \quad} The fixed-size input-output design allows CP-DRL to leverage multi-task training (MT) with varying machine numbers and resources. We refer to this multi-task extension as CP-DRL(MT). We trained the CP-DRL(MT)
with $N \in \{2, 3, 4, 5\}$, randomly switching configurations at the end of each episode over 2000 training episodes. CP-DRL(MT) was evaluated separately, and GGF values for WIP and RDM policies were obtained from 1000 Monte Carlo runs. The numbers following the plus-minus sign ($\pm$) represent the variance across 5 experiments with different random seeds in Table \ref{tab:performance1} and \ref{tab:performance2}. Variances for WIP and RDM are minimal and omitted, with bold font indicating the best GGF scores at each row excluding optimal values. As shown in Table \ref{tab:performance1}, CP-DRL(MT) consistently achieves scores very close to the OPT values as the number of machines increases from 2 to 4. For the 5-machine case, CP-DRL(MT) shows slightly better performance than the single-task CP-DRL. In Table \ref{tab:performance2}, the single- and multi-task CP-DRL agents show slight variations in performance across different machine numbers. For $N=5$, CP-DRL achieves the best GGF score, slightly outperforming WIP. {In both cases, the CP-DRL approach outperforms Vanilla-DRL, the three heuristic methods, and the random agent.}

\begin{table*}[!htb] 
\centering
\caption{GGF Scores (Exponential-RCCC)}
\label{tab:performance1}
\begin{tabular}{ccccccccc>{\centering\arraybackslash}c}
        \hline
        $N$ & OPT & WIP & CP-DRL & CP-DRL(MT)  & V-DRL& OFT & MYP & EQR & RDM \\ \hline
        2 & 14.19 & 14.07 & \textbf{14.12} $\pm$ 0.01 & 14.11 $\pm$ 0.01  & 13.56 
$\pm$ 0.00&5.84& 12.59&10.05& 9.67 \\
        3 & 14.08 & 13.75 & \textbf{13.95} $\pm$ 0.02 & 13.89 $\pm$ 0.14  &  13.39 $\pm$ 0.00&7.92& 12.32&	11.67& 10.13 \\
        4 & 13.94 & 13.27 & \textbf{13.64} $\pm$ 0.05 & 13.59 $\pm$ 0.10  &  13.04 $\pm$ 0.01&9.02& 12.86&12.03& 9.74 \\
        5 & 13.77 & 12.47 & 12.96 $\pm$ 0.01 & \textbf{13.28} $\pm$ 0.03  &  12.83 $\pm$ 0.00&10.01& 12.08&11.87& 8.95 \\ \hline
        \end{tabular}
\end{table*}

\begin{table*}[!htb]
\centering
\caption{GGF Scores (Quadratic-RCCC)}
\label{tab:performance2}
\begin{tabular}{ccccccccc>{\centering\arraybackslash}c}
\hline
$N$ & OPT & WIP & CP-DRL & CP-DRL(MT)  & V-DRL& OFT & MYP & EQR & RDM \\ \hline
2 & 16.17 & \textbf{16.17} & 16.14 $\pm$ 0.00 & 16.14 $\pm$ 0.00  & 15.36 
 $\pm$ 0.00& 3.11& 6.61&9.73& 10.15 \\
3 & 16.10 & \textbf{16.09} & 16.05 $\pm$ 0.00 & 16.05 $\pm$ 0.00  & 15.17 $\pm$ 0.01& 6.16& 6.63&12.73& 11.83 \\
4 & 16.01 & \textbf{16.01} & 15.94 $\pm$ 0.00 & 15.94 $\pm$ 0.00  & 15.01 $\pm$ 0.00	& 7.92& 6.85&14.02& 12.17 \\
5 & 15.91 & 15.86 & \textbf{15.87} $\pm$ 0.02 & 15.86 $\pm$ 0.02  & 14.73 $\pm$ 0.00& 9.25& 6.70&14.64& 11.98 \\ \hline
\end{tabular}
\end{table*}

\begin{figure*}[!htb] 
    \centering
    \begin{minipage}{0.325\textwidth}
        \centering
        \includegraphics[width=\linewidth]{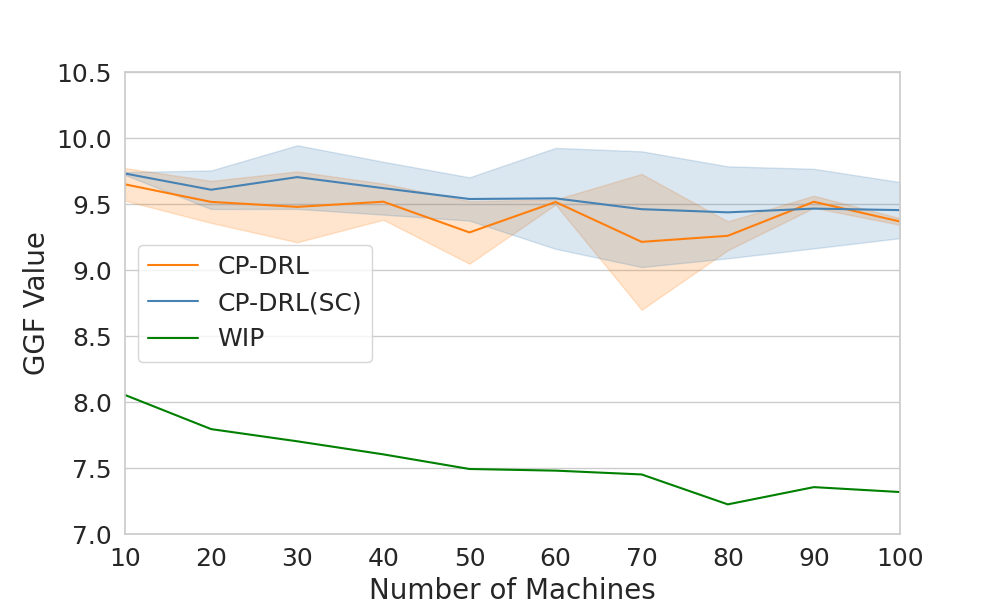}
        \caption*{\small{(a) GGF values for the number of machines $N \in [10, 100]$}}
    \end{minipage}
    \hspace{-1mm}
    \begin{minipage}{0.325\textwidth}
        \centering
        \includegraphics[width=\linewidth]{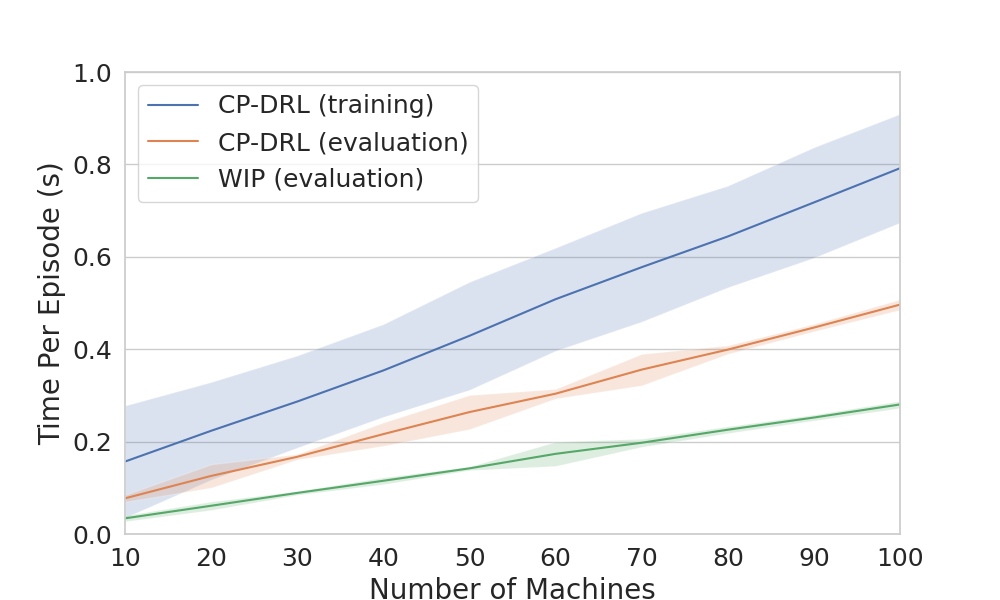}
        \caption*{\small{(b) Time per episode in seconds with a resource ratio $b/N =0.1$}}
    \end{minipage}
    \hspace{-1mm}
    \begin{minipage}{0.325\textwidth}
        \centering
        \includegraphics[width=\linewidth]{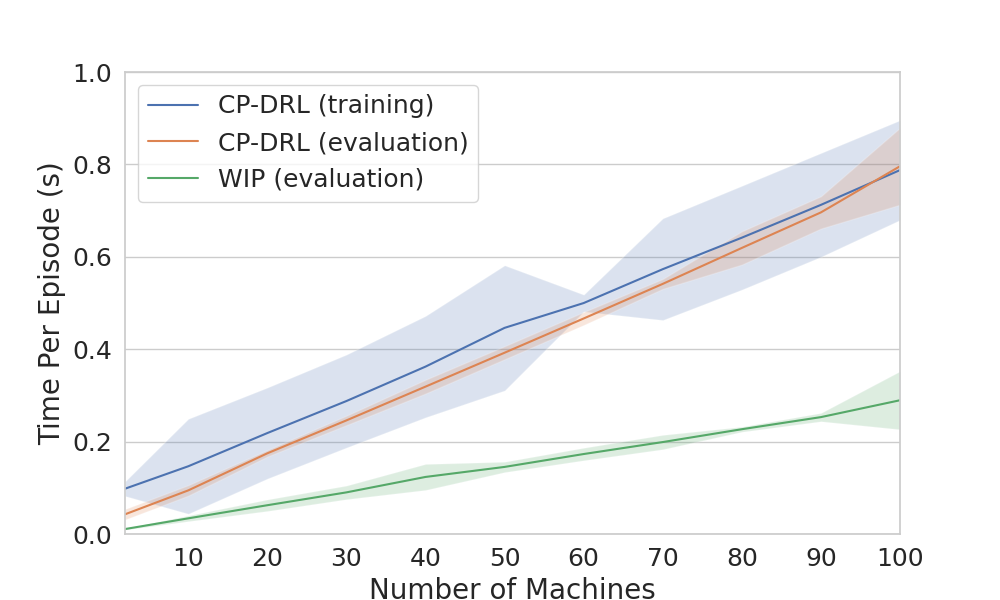}
        \caption*{\small{(c) Time per episode in seconds with a resource ratio $b/N =0.5$}}
    \end{minipage}
    
    \caption{
    {\textbf{(Colored) Scalability and Time Efficiency of CP-DRL.} \quad Subfigures (a) and (b) show the scalability of CP-DRL with a fixed resource ratio of 0.1. Subfigure (a) presents GGF values across different machine counts, with intervals representing the standard deviation over 5 runs. Subfigure (b) and (c) depicts time per episode in seconds for a fixed resource ratio of 0.1 and 0.5, respectively. In all time plots, the green line represents WIP during MC evaluation, the blue line shows CP-DRL during training, and the orange line represents CP-DRL during MC evaluation.}}
    \label{fig:combined}
\end{figure*}

\textbf{Experiment 3 (Scalability) \quad} We assess CP-DRL scalability by increasing the number of machines while keeping the resource proportion at 0.1 for the \textit{Exponential-RCCC} instances. We refer to this scaled extension as CP-DRL(SC). We vary the number of machines from 10 to 100 to evaluate CP-DRL performance as the problem size grows. We also use CP-DRL(SC), trained on 10 machines with 1 unit of resource, and scale it to tasks with 20 to 100 machines. Figure \ref{fig:combined}a shows CP-DRL and CP-DRL(SC) consistently achieve higher GGF values than WIP as machine numbers increase. CP-DRL(SC) delivers results comparable to separately trained CP-DRL, reducing training time while maintaining similar performance. Both WIP and CP-DRL show linear growth in time consumption per episode as machine numbers scale up.

\textbf{Experiment 4 (Efficiency) \quad} In the GGF-LP model (\ref{ggf-mdp-d}), the number of constraints grows exponentially with the number of machines $N$ as $N^2 + S^N$, and the variables increase by $2N + (N+1) \cdot S^N$. Using the count dual LP model (\ref{eq:count-dlp}) reduces the model size, but constraints still grow as $\binom{N+S-1}{S-1}$ and variables increase by $\binom{N+S-1}{S-1} \cdot A$. These growth patterns create computational challenges as the problem size increases. A detailed time analysis for the GGF-LP and count dual LP models with $N$ from 2 to 7 is provided in Appendix \ref{appendix:solving-time}.
In addition to the time per episode for a fixed ratio of 0.1 in Figure \ref{fig:combined}a, we analyze performance with a 0.5 ratio (Figure \ref{fig:combined}c) and varying machine proportions, keeping the number of machines fixed at 10. We evaluate CP-DRL over machine proportions from 0.1 to 0.9. The results show that the time per episode increases linearly with the number of machines, while training and evaluation times remain relatively stable. This indicates that the sampling procedure for legal actions is the primary bottleneck. Meanwhile, the resource ratio has minimal impact on computing times.

\section{CONCLUSION}
We incorporate the fairness consideration in terms of the generalized Gini function within the weakly coupled Markov decision processes, and define the GGF-WCMDP optimization problem. First, we present an exact method based on linear programming for solving it. We then derive an equivalent problem based on a utilitarian reduction when the WCMDP is symmetric, and show that the set of optimal permutation invariant policy for the utilitarian objective is also optimal for the original GGF-WCMDP problem. We further leverage this result by utilizing a count state representation and introduce a count-proportion-based deep RL approach to devise more efficient and scalable solutions. Our empirical results show that the proposed method using PPO as the RL algorithm consistently achieves high-quality GGF solutions. Moreover, the flexibility provided by the count-proportion approach offers possibilities for scaling up to more complex tasks and context where Whittle index policies are unavailable due to the violation of the indexability property by the sub-MDPs.

\section*{Acknowledgements}
Xiaohui Tu was partially funded by GERAD and IVADO.
Yossiri Adulyasak was partially supported by the Canadian Natural Sciences and Engineering Research Council [Grant RGPIN-2021-03264] and by the Canada Research Chair program [CRC-2022-00087].
Nima Akbarzadeh was partially funded by GERAD, FRQNT, and IVADO.
Erick Delage was partially supported by the Canadian Natural Sciences and Engineering Research Council [Grant RGPIN-2022-05261] and by the Canada Research Chair program [950-230057].

\bibliographystyle{apalike}
\bibliography{reference}

\newpage
\onecolumn
 
\appendix
\section{RELATED WORK} \label{appendix:literature}
Fairness-aware learning is increasingly integrated into the decision-making ecosystem to accommodate minority interests. However, naively imposing fairness constraints can actually exacerbate inequity \citep{wen2021algorithms} if the feedback effects of decisions are ignored. Many real-world fairness applications are not one-time static decisions \citep{zhao2019inherent} and can thus be better modeled with sequential decision problems, which still remain relatively understudied. 

{\textbf{Fairness in resource allocation}\quad Fairness has been an important concern in resource allocation problems, where traditional approaches often build upon optimization frameworks, leveraging fairness-constrained optimization \citep{argyris2022fair, xinying2023guide}, game-theoretic concepts \citep{namazi2023novel}, or axiomatic principles \citep{lan2011axiomatic} to derive fair solutions. These models provide practical fair solutions, but do not account for the long-term impact of allocation decisions, which motivates fair-aware sequential decision-making frameworks based on Markov decision processes (MDPs) as discussed next.}

\textbf{Fairness with dynamics}\quad There are a few studies investigating fairness-aware sequential decision making {without relying on MDPs}. For instance, \cite{liu2018delayed} consider one-step delayed feedback effects, \cite{creager2020causal} propose causal modeling of dynamical systems to address fairness, and \cite{zhang2019group} construct a user participation dynamics model where individuals respond to perceived decisions by leaving the system uniformly at random. 
These studies extend the fairness definition in temporally extended decision-making settings, but do not take feedback and learning into account that the system may fail to adapt to changing conditions.
\cite{alamdari2023remembering} address this gap by introducing multi-stakeholder fairness as non-Markovian sequential decision making and developing a Q-learning based algorithm with counterfactual experiences to enhance sample-efficient fair policy learning.

\textbf{Fairness in Markov decision processes}\quad
\cite{zhang2020fair} consider how algorithmic decisions impact the evolution of feature space of the underlying population modeled as MDPs but is limited to binary decisions.
\cite{ghalme2022long} study a fair resource allocation problem in the average MDP setting and proposes an approximate algorithm to compute the policy with sample complexity bounds. 
However, their definition of fairness is restricted to the minimum visitation frequency across all states, potentially resulting in unbalanced rewards among {sub-MDPs}.
\cite{wen2021algorithms} develop fair decision-making policies in discounted MDPs, but the performance guarantees are achieved only under a loose condition. In contrast, our work takes into account a more comprehensive definition of fairness. \cite{segal2023} investigate the impact of societal bias dynamics on long-term fairness and the interplay between utility and fairness under various optimization parameters. Additionally, \cite{hassanzadeh2023sequential} address a fair resource allocation problem similar to our work but in continuous state and action space. They define fairness to the agents considering all their allocations over the horizon under the Nash Social Welfare objective in hindsight.

\textbf{Fairness in reinforcement learning}\quad \cite{jabbari2017fairness} initiate the \textit{meritocratic fairness} notion from the multi-arm bandits setting to the reinforcement learning (RL) setting. Later, fairness consideration has been integrated in RL to achieve fair solutions in different domains, including a fair vaccine allocation policy that equalizes outcomes in the population \citep{atwood2019fair}, balancing between fairness and accuracy for interactive user recommendation \citep{liu2020balancing, ge2022toward}, and fair IoT that  continuously monitors the human state and changes in the environment to adapt its behavior accordingly \citep{elmalaki2021fair}. However, most work focuses on the impartiality aspect of fairness. \cite{jiang2019learning} investigate multi-agent RL where fairness is defined over agents and encoded with a different welfare function, but the focus is on learning decentralized policies in a distributed way.
We refer readers to two literature review papers by \cite{gajane2022survey} and \cite{ reuel2024fairsurvey}
on fairness considerations in RL, which provide comprehensive insights into current trends, challenges, and methodologies in the field.

\textbf{Fairness in restless multi-arm bandits}\quad A line of work closely related to ours focuses on fairness in restless multi-arm bandits (RMABs).
\cite{li2022efficient} first introduce the consideration of fairness in restless bandits by proposing an algorithm that ensures a minimum number of selections for each arm. 
Subsequent studies have explored similar individual fairness constraints, which aim to distribute resources equitably among arms but in a probabilistic manner. For instance, \cite{herlihy2023planning} introduce a method that imposes a strictly positive lower bound on the probability of each arm being pulled at each timestep. \cite{li2022towards, li2023avoiding} investigate fairness by always probabilistically favoring arms that yield higher long-term cumulative rewards. Additionally, \cite{sood2024fairness} propose an approach where each arm receives pulls in proportion to its merit, which is determined by its stationary reward distribution. Our work differs from these approaches by explicitly aiming to prevent disparity and ensure a more balanced reward distribution among all arms through the generalized Gini welfare objective. The only work that considers the Gini index objective is by \cite{vermagroup}, which develops a decision-focused learning pipeline to solve equitable RMABs. In contrast, our work applies to a more general setting on weakly coupled MDPs, and does not rely on the Whittle indexability of the coupled MDPs.

\section{PROOFS OF SECTION \ref{sec:reduction}}

We start this section with some preliminary results regarding 1) the effect of replacing a policy with one that has permuted indices on the value function of a symmetric WCMDP (Section \ref{sec:valueFunderPermPolicy}); and 2) a well-known result from \cite{puterman2014markov} on the equivalency between stationary policies and occupancy measures (Section \ref{sec:puterman}). This is followed by the proof of Lemma \ref{thm:bar-policy}, which helps establish our main result, Theorem \ref{thm:eqv} in Section \ref{A:eqv}.

\subsection{Value Function under Permuted Policy for Symmetric WCMDP}\label{sec:valueFunderPermPolicy}

\begin{lemma}
\label{thm:piQVQ}
If a weakly coupled WCMDP is symmetric (definition \ref{def:s-WCMDP}), then for any policy $\vpi$ and permutation operator $Q$, we have 
$\Vm = Q \V^{\ppi}_0,$
where the permuted policy $\ppi(\s,\a):=\vpi(Q\s,Q\a)$ for all $(\s,\a)$ pairs.
\end{lemma}
This lemma implies an important equivalency in symmetric weakly coupled MDPs with identical sub-MDPs. If we permute the states and actions of a policy, the permuted version of the resulting value function is equivalent to the original value function.

\textbf{Proof:}
We can first show that for all $t$,
\[\Prob^{\ppi}(\s_t=\s,\a_t=\a|\s_0=\so) = \Prob^{\vpi}(\s_t=Q \s,\a_t = Q \a|\s_0 = Q \so).\]
This can be done inductively. Starting at $t=0$, we have that:
\begin{align*}
\Prob^{\ppi}(\s_0=\s,\a_0=\a|\s_0=\so)
&=\ppi(\s,\a)\sI\{\s=\so\}\\
&= \vpi(Q\s,Q\a)\sI\{Q\s=Q\so\}\\
&=\Prob^{\vpi}(\s_0=Q \s,\a_0=Q \a| \s_0=Q\so).
\end{align*}
Next, assuming that $\Prob^{\ppi}(\s_t=\s,\a_t=\a|\s_0=\so) = \Prob^{\vpi}(\s_t=Q \s,\a_t = Q \a|\s_0= Q \so)$, we can show that it is also the case for $t+1$:
\begin{align*}
&\Prob^{\ppi}(\s_{t+1}=\s',\a_{t+1}=\a'|\s_0=\so) \\
&= \ppi(\s',\a')\sum_{\s,\a}  \Prob(\s_{t+1}=\s'|\s_t=\s,\a_{t}=\a) \Prob^{\ppi}(\s_t=\s,\a_t=\a|\s_0=\so)\\
&= \vpi(Q \s',Q \a')\sum_{\s,\a}  \Pn(\s'|\s,\a) \Prob^{\pi}(\s_t=Q \s,\a_t = Q \a|\s_0= Q \so)\\
&= \vpi(Q \s',Q \a')\sum_{\s,\a}  \Pn(Q \s'|Q \s,Q \a) \Prob^{\pi}(\s_t=Q \s,\a_t=Q \a|\s_0=Q \so)\\
&=\Prob^{\vpi}(\s_{t+1}=Q \s',\a_{t+1}=Q \a'|\s_0=Q\so),
\end{align*}
where we use the fact that the sub-MDPs are identical so that $\Pn(\s'|\s,\a)=\Pn(Q\s'|Q\s,Q\a)$.

We now have that, 
\begingroup\allowdisplaybreaks
\begin{align*}
    \V^{\ppi}_0 &= \sum_{\s,\a}\sum_{\so}\vmu(\so)\sum_{t=0}^\infty \gamma^t \Prob^{\ppi}(\s_t=\s,\a_t=\a|\s_0=\so) \r(\s,\a)\\
    &= \sum_{\s,\a}\sum_{\so}\vmu(\so)\sum_{t=0}^\infty \gamma^t \Prob^{\vpi}(\s_t= Q \s,\a_t= Q\a|\s_0=Q\so) \r(\s,\a)\\
    &= \sum_{\s,\a}\sum_{\so}\vmu(Q \so)\sum_{t=0}^\infty \gamma^t \Prob^{\vpi}(\s_t= Q \s,\a_t= Qa|\s_0=Q\so) \r(\s,\a)\\
    &= \sum_{\s,\a}\sum_{\so}\vmu(Q \so)\sum_{t=0}^\infty \gamma^t \Prob^{\vpi}(\s_t= Q \s,\a_t= Q\a|\s_0=Q\so) Q^{-1}\r(Q\s,Q\a)\\
    &= Q^{-1} \left(\sum_{\s,\a}\sum_{\so}\vmu(Q \so)\sum_{t=0}^\infty \gamma^t \Prob^{\vpi}(\s_t= Q \s,\a_t= Q\a|\s_0=Q\so) \r(Q\s,Q\a)\right)\\
    &= Q^{-1} \left(\sum_{\s',\a'}\sum_{\so'}\vmu(\so')\sum_{t=0}^\infty \gamma^t \Prob^{\vpi}(\s_t= \s',\a_t= \a'|\s_0=\so') \r(\s',\a')\right) \;=\; Q^{-1} \Vm,
\end{align*}
\endgroup
where we first use the relation between $P^{\ppi}$ and $\vpi$, then exploit the permutation invariance of $\vmu$. We then exploit the permutation invariance $Q \r(\s,\a)= \r(Q\s, Q\a)$, and reindex the summations using $\s':=Q \s$, $\a':=Q \a$, and $\so':=Q\so$. \hfill $\square$

\subsection{Mapping Between Stationary Policies and Occupancy Measures} \label{sec:puterman}
We present results of Theorem 6.9.1 in \cite{puterman2014markov} to support Lemma \ref{thm:bar-policy}. A detailed proof is provided in the book.

\begin{lemma} (Theorem 6.9.1 of \cite{puterman2014markov})
\label{prop:x-pi-x}
    Let $\Pi$ denote the set of stationary stochastic Markov policies and $\mathcal{X}$ the set of occupancy measures. There exists a bijection $h: \Pi \rightarrow \gX$ such that for any policy $\vpi$, $h(\vpi)$ uniquely corresponds to its occupancy measure $q_{\vpi}$. Specifically, there is a one-to-one mapping between policies and occupancy measures satisfying:
    \begin{enumerate}[leftmargin=*]
        \item For any policy $\vpi \in \Pi$, the occupancy measure $q_{\vpi}: \Sn \times \An \rightarrow \sR$ is defined as
            \begin{equation}
                q_{\vpi}(\s, \a) := \sum_{\so \in \Sn} \vmu(\so) \sum_{t=0}^{\infty} \gamma^{t} \Prob^{\vpi}\left(\s_t=\s, \a_t=\a | \s_0=\so \right),
            \label{eq:x-pi}
            \end{equation}
        for all $\a \in \An$ and $\s \in \Sn$.
        \item For any occupancy measure $q(\s, \a): \Sn \times \An \rightarrow \sR$, the policy $\vpi_{q}$ is constructed as
            \begin{equation}
                \vpi_{q}(\s, \a):=\frac{q(\s, \a)}{\sum\limits_{\a^{\prime} \in \An} q\left(\s, \a^{\prime}\right)},
            \label{eq:pi-x}
            \end{equation}
        for all $\a \in \An$ and $\s \in \Sn$. 
    \end{enumerate}
    It follows that $\vpi = \vpi_{q_\pi}$.
\end{lemma}

Now, we show that the value function can be represented using occupancy measures.

\begin{lemma} For any policy $\vpi \in \Pi$, and the occupancy measure $q_{\vpi}$ defined by (\ref{eq:x-pi}), the expected total discounted rewards under the policy $\vpi$ can be expressed as:
\begin{equation}\label{eq:Vm}
    \Vm = \sum_{\s \in \Sn} \sum_{\a \in \An} q_{\vpi}(\s, \a) \r(\s, \a).
\end{equation}
\label{prop:xr=muv}
\end{lemma}
\textbf{Proof.} Expanding the expected total discounted rewards $\Vm$ (as defined by \eqref{eq:V}), we have:
\begin{equation*}
    \Vm = \sum_{\so \in \Sn} \vmu(\so) \sum_{\s \in \Sn} \sum_{\a \in \An} \sum_{t=0}^{\infty} \Prob^{\vpi}\left(\s_t=\s, \a_t=\a | \s_0=\so \right) \gamma^t \r(\s, \a).
\end{equation*}
Rearranging the terms:
\begin{equation*}
    \Vm = \sum_{\s \in \Sn} \sum_{\a \in \An} \left(\sum_{\so \in \Sn} \vmu(\so) \sum_{t=0}^{\infty} \gamma^t \Prob^{\vpi}\left(\s_t=\s, \a_t=\a | \s_0=\so \right) \right) \r(\s, \a).
\end{equation*}
Replacing the term in parentheses as the occupancy measure $q_{\vpi}(\s, \a)$ in (\ref{eq:x-pi}) leads directly to \eqref{eq:Vm}, which completes the proof. \hfill $\square$

\newpage
\subsection{Proof of Lemma \ref{thm:bar-policy}} \label{A:bar-policy}

\textbf{Lemma \ref{thm:bar-policy} (Uniform State-Value Representation)}
\textit{If a WCMDP is symmetric (definition \ref{def:s-WCMDP}), then for any policy $\vpi$, there exists a corresponding permutation invariant policy $\bpi$ such that the vector of expected total discounted rewards for all sub-MDPs under $\bpi$ is equal to the average of the expected total discounted rewards for each sub-MDP, i.e., $$\V^{\bpi}_0
= \frac{1}{N} \sum_{n=1}^N V_{0,n}^\vpi
\vone.$$}

\textbf{Proof by construction.} We first construct, for any fixed $Q$, the permuted policy $\ppi(\s,\a):=\vpi(Q \s, Q \a)$ and characterize its occupancy measure $q_{\ppi}$ as
\begin{equation}
    q_{\ppi}(\s, \a) := \sum_{\so \in \Sn} \vmu(\so) \sum_{t=0}^{\infty} \gamma^{t} \Prob^{\ppi}\left(\s_t=\s, \a_t=\a | \s_0=\so\right).
    \label{eq:xq}
\end{equation}
Next, we construct a new measure $\bar{q}$ obtained by averaging all permuted occupancy measures $q_{\ppi}$ for $Q \in \G$ on all $(\s, \a)$ pairs as
\begin{equation}
    \bar{q}(\s, \a) := \frac{1}{N!} \sum_{Q} q_{\ppi}(\s, \a).
    \label{eq:bar-pi-x}
\end{equation}
One can confirm that $\bar{q}$ is an occupancy measure, i.e., $\bar{q}\in\mathcal{X}$, since each $q_\ppi\in\mathcal{X}$ and $\mathcal{X}$ is convex. Indeed, the convexity of $\mathcal{X}$ easily follows from that the fact that it contains any measure that it is the set of measures that satisfy constraints \ref{eq:ggf-mdp-d:c2} and \ref{eq:ggf-mdp-d:c3}.

From Lemma \ref{prop:x-pi-x}, a stationary policy $\bpi$ can be constructed such that its occupancy measure matches $\bar{q}(\s, \a)$:
\[q_{\bpi}(\s, \a):=\sum_{\so \in \Sn} \mu(\so) \sum_{t=0}^{\infty} \gamma^{t} \Prob^{\bpi}\left(\s_t=\s, \a_t=\a | \s_0=\so\right)= \bar{q}(\s, \a),\forall \s,\a.\]

We can then derive the following steps:

\begingroup\allowdisplaybreaks
\begin{align*}
    \V^{\bpi}_0 &=\sum_{\s \in \Sn} \sum_{\a \in \An} q_{\bpi}(\s, \a) \r(\s, \a) \quad  \mbox{(By Lemma \ref{prop:xr=muv})}\\
    &=\sum_{\s \in \Sn} \sum_{\a \in \An} \bar{q}(\s, \a) \r(\s, \a) \quad  \mbox{(By Lemma \ref{prop:x-pi-x})}\\
    &=\frac{1}{N!} \sum_{Q \in \G}
    \sum_{\s \in \Sn} \sum_{\a \in \An} q_{\ppi}(\s, \a) \r(\s, \a) \quad  \mbox{(By construction in \eqref{eq:bar-pi-x}) }\\
    &= \frac{1}{N!} \sum_{Q \in \G}  \Vo^{\ppi} \quad  \mbox{(By Lemma \ref{prop:xr=muv})}\\
    &= \frac{1}{N!} \sum_{Q \in \G} Q^{-1} \Vm \quad \mbox{(By Lemma \ref{thm:piQVQ})}\\
    &= \frac{1}{N!} \sum_{Q \in \G} Q^{-1} \begin{bmatrix}
    V_{0,1}^{\vpi} \\ \cdots \\ V_{0,N}^{\vpi} \\\end{bmatrix} \quad \text{(Vector form)} \\
    &= \frac{1}{N!} \begin{bmatrix}
    (N-1)! \sum_{n=1}^N V_{0, n}^{\vpi} \\ \cdots \\ (N-1)! \sum_{n=1}^N V_{0,n}^{\vpi}\end{bmatrix} \quad \text{(Property of permutation group)}\\
    &= \frac{1}{N!} (N-1)! \sum_{n=1}^N V_{0,n}^{\vpi} \vone\\
    &= \frac{1}{N} \sum_{n=1}^N V_{0,n}^{\vpi} \vone.
\end{align*}
\endgroup

\newpage
We complete this proof by demonstrating that $\bpi$ is permutation invariant. Namely, for all $Q\in\G$, we can show that:
\begin{equation*}
\begin{aligned}
\bpi(Q\s, Q\a) &\propto \frac{1}{N!} \sum_{Q'\in\G} q_{\vpi^{Q'}}\left(Q\s, Q\a\right) \\
&= \frac{1}{N!} \sum_{Q'\in\G} q_{\vpi}(Q'Q\s, Q'Q\a) \\
&= \frac{1}{N!} \sum_{Q''\in\G} q_{\vpi}(Q''\s, Q''\a) \\
&= \frac{1}{N!} \sum_{Q''\in\G} q_{\vpi^{Q''}}(\s, \a) \\
&\propto \bpi(\s, \a).
\end{aligned}
\end{equation*}
\hfill $\square$

\subsection{Proof of Theorem \ref{thm:eqv}}\label{A:eqv}

We start with a simple lemma.

\begin{lemma}\label{lemma:weights}
    For any $\vw$ and any $\vv\in\mathbb{R}^N$, we have that $GGF_{\voneN}[\vv]\geq \GGF_{\vw}[\vv]$.
\end{lemma}

\textbf{Proof.} This simply follows from:
\begin{align*}
    \GGF_{\voneN}[\vv] &= \frac{1}{N}\vone^\top \vv = \left(\frac{1}{N!}\sum_{Q\in\G} Q\vw\right)^\top \vv \geq \min_{Q\in\G} (Q\vw)^\top \vv = \GGF_{\vw}[\vv].
\end{align*}
\hfill $\square$

\textbf{Theorem \ref{thm:eqv} (Utilitarian Reduction)}
\textit{For a symmetric WCMDP (definition \ref{def:s-WCMDP}), let $\Pi^*_{{\voneN,\mymbox{PI}}}$ be the set of optimal policies for the utilitarian approach (definition \ref{def:utilitarian}) that is permutation invariant, then $\Pi^*_{\voneN,\mymbox{PI}}$ is necessarily non-empty and all $\pi^*_{\voneN,\mymbox{PI}}\in\Pi^*_{\voneN,\mymbox{PI}}$ satisfies }
\begin{equation*}
\GGF_{\vw}[\V_0^{\pi_{\voneN,\mbox{\tiny{PI}}}^{*}}]
= \max_{\vpi} \GGF_{\vw}\left[\Vm \right], \, \forall \vw \in\Delta(N).
\end{equation*}

\textbf{Proof.} Let us denote an optimal policy to the special case of the GGF-WCMDP problem (\ref{eq:ggf}) with equal weights as $\vpi^*_{\voneN}$:
\begin{equation}\label{eq:pi-ew}
\vpi^*_{\voneN} \in \arg \max_{\vpi} \GGF_{\voneN}\left[\Vm\right].
\end{equation}
Based on Lemma \ref{thm:bar-policy}, we can construct a permutation invariant policy
$\bpi^*_{\voneN}$ satisfying
\begin{equation}
\bV^{\vpi^*_{\voneN}}
\vone = \V_0^{\bpi^*_{\voneN}},
\label{eq:bar-policy-1}
\end{equation}
then with equation (\ref{eq:bar-policy-1}) and the fact that any weight vector $\vw$ must sum to 1, we have that
\[
\GGF_{\voneN}\left[\V_0^{\vpi^*_{\voneN}}\right] 
= \bV^{\vpi^*_{\voneN}}
= \GGF_{\vw}\left[\frac{1}{N} \sum_{n=1}^{N} V_{0,n}^{\vpi^*_{\voneN}}
\vone\right]=\GGF_{\vw}\left[\V_0^{\bpi^*_{\voneN}}\right] \;,\;\forall \vw.
\]

Furthermore, given any $\vw$, let us denote with  $\vpi_{\vw}^*$ any optimal policy to the GGF problem with $\vw$ weights. One can establish that:
\begin{align}\label{ineq:ggf-weights-1}
\GGF_{\vw}[\V_0^{{\vpi}_{\vw}^*}] \geq \GGF_{\vw}\left[\V_0^{\bpi^*_{\voneN}}\right] = \GGF_{\voneN}\left[\V_0^{\vpi^*_{\voneN}}\right] \geq \GGF_{\voneN}\left[\V_0^{{\vpi}_{\vw}^*}\right].
\end{align}

Considering that the largest optimal value for the GGF problem is achieved when weights are equal (see Lemma \ref{lemma:weights}):
\begin{align*}
\GGF_{\voneN}\left[\V_0^{{\vpi}}\right]\geq \GGF_{\vw}[\V_0^{{\vpi}}] 
, \forall \vpi,\;\forall\vw\in\Delta(N).
\end{align*}

The inequalities in (\ref{ineq:ggf-weights-1}) should therefore all reach equality:
\begin{align*}
\GGF_{\vw}[\V_0^{\vpi_w^*}] = \GGF_{\vw}\left[\V_0^{\bpi^*_{\voneN}}\right] = \GGF_{\voneN}\left[\V_0^{\vpi^*_{\voneN}}\right].
\end{align*}
This implies that the bar optimal policy constructed from any optimal policy to the utilitarian approach remains optimal for any weights in the GGF optimization problem.
{Furthermore, it implies that there exists at least one permutation invariant policy that is optimal for the utilitarian approach. 

Now, let us take any optimal permutation invariant policy $\pi_{\voneN,\mymbox{PI}}^*$ to the utilitarian problem. The arguments above can straightforwardly be reused to get the conclusion that $\bar{\pi}_{\voneN,\mymbox{PI}}^*$ have the same properties as the originally constructed $\bar{\pi}$. Namely, for all $\vw$,
\begin{align*}
\GGF_{\vw}[\V_0^{\vpi_w^*}] = \GGF_{\vw}\left[\V_0^{\bpi^*_{\voneN,\mymbox{\tiny{PI}}}}\right] = \GGF_{\voneN}\left[\V_0^{\vpi^*_{\voneN}}\right],\;\forall \vw\in\Delta(N). 
\end{align*}
Looking more closely at $\bpi^*_{\voneN,\mymbox{PI}}$, we observe that for any $\s$ and $\a$:
\begin{align*}
    \bpi^*_{\voneN,\mymbox{PI}}(\s,\a)&\propto \frac{1}{N!} \sum_{Q'\in\G} q_{\pi_{\voneN,\mymbox{\tiny{PI}}}^{*,Q'}}\left(\s, \a\right)\\
    &=\frac{1}{N!} \sum_{Q'\in\G} q_{\pi_{\voneN,\mymbox{\tiny{PI}}}^{*}}\left(\s, \a\right)\\
    &=q_{\pi_{\voneN,\mymbox{\tiny{PI}}}^{*}}\left(\s, \a\right)\\
    &\propto \pi^{*}_{\voneN,\mymbox{{PI}}}(\s,\a).
\end{align*}
Hence, we have that $\bpi^*_{\voneN,\mymbox{PI}}=\pi^*_{\voneN,\mymbox{PI}}$. This thus implies that the permutation invariant $\pi_{\voneN,\mymbox{PI}}^*$ already satisfied these properties, i.e.,
\begin{align*}
\GGF_{\vw}[\V_0^{\vpi_w^*}] = \GGF_{\vw}\left[\V_0^{\pi^*_{\voneN,\mymbox{\tiny{PI}}}}\right] = \GGF_{\voneN}\left[\V_0^{\vpi^*_{\voneN}}\right],\;\forall \vw\in\Delta(N). 
\end{align*}
}

\hfill $\square$

\subsection{{Extension to Other Fairness Measures}} \label{appendix:extension}

The utilitarian reduction (Theorem \ref{thm:eqv}) can be extended to a broader scope of fairness measures{, where we replace the GGF measure in optimization problem (\ref{eq:ggf}) and define the $\rho$-WCMDP problem accordingly.}

\newcommand{\mcV}{\mathcal{V}}
\begin{corollary}\label{cr:utilitarian}
Let $\rho:\mcV\rightarrow\mathbb{R}$, with $\mcV\subseteq\mathbb{R}^N$, be a fairness measure  that satisfies:
\begin{itemize}
    \item (Concavity): The set $\mcV$ is convex and $\forall \vv,\vw\in\mcV$, and $\theta\in[0,1]$, $\rho[\theta \vv + (1-\theta)\vw]\geq \theta\rho[\vv]+(1-\theta)\rho[\vw]$
    \item (Permutation invariance): $\forall \vv\in\mcV$ and all $Q\in\G$, both $Q\vv\in\mcV$ and $\rho[\vv]=\rho[Q \vv]$
    \item (Constant vector invariant) $\forall \bar{v} \in\mathbb{R}\cap\mcV$, $\rho[\bar{v} \vone]= \bar{v}$.
\end{itemize}
For a symmetric WCMDP such that $\V_0^\pi\in\mcV$ for all $\pi$, let $\Pi^*_{\mymbox{U, PI}}$ be the set of optimal policies for the utilitarian approach that is permutation invariant, then $\Pi^*_{\mymbox{U, PI}}$ is necessarily non-empty and all $\pi^*_{\mymbox{U, PI}}\in\Pi^*_{\mymbox{U, PI}}$ satisfies 
\[
\rho[\V_0^{\pi_{\mbox{\tiny{U, PI}}}^{*}}]
= \max_{\vpi} \rho\left[\Vm \right].
\]

\end{corollary}

\textbf{Proof.} Similar to the proof of Theorem \ref{thm:eqv}, we define the utilitarian fairness measure as
$$\rho_{\ut}[\vv] = \frac{1}{N}\sum_{n=1}^N v_n.
$$
Defining the optimal policy to the $\rho_{\ut}$-WCMDP problem with utilitarian objective as $\vpi^*_{\ut} \in \arg \max\limits_{\vpi} \rho_{\ut}[\Vm]$ and with Lemma \ref{thm:bar-policy}, we can construct a permutation invariant policy $\bar{\pi}^*_{\ut}$ satisfying
\begin{equation}
\bV^{\vpi^*_{\ut}}
\vone = \V_0^{\bpi^*_{\ut}}\in\mcV,
\end{equation}
with $\bV^{\vpi^*_{\ut}}:=\frac{1}{N}\sum_{n=1}^N V_{0,n}^{\vpi^*_{\ut}}$.
Then we have that
\[
\rho_{\ut}\left[\V_0^{\vpi^*_{\ut}}\right] 
= \bV^{\vpi^*_{\ut}}
= \rho\left[\bV^{\vpi^*_{\ut}}\right]=\rho\left[\frac{1}{N} \sum_{n=1}^{N} V_{0,n}^{\vpi^*_{\ut}}
\vone\right]=\rho\left[\V_0^{\bpi^*_{\ut}}\right],
\]
where we exploit $\rho[\bar{v}\vone]=\bar{v}$ for all $\bar{v} \in\mathbb{R}\cap\mcV$.
Furthermore, let $\vpi^*$ be any optimal policy to the $\rho$-WCMDP problem. One can establish that:    
\begin{align}\label{ineq:ggf-extension}
\rho[\V_0^{{\vpi}^*}] \geq \rho\left[\V_0^{\bpi^*_{\ut}}\right] = \rho_{\ut}\left[\V_0^{\vpi^*_{\ut}}\right] \geq \rho_{\ut}\left[\V_0^{{\vpi}^*}\right].
\end{align}
By Jensen's inequality and the fact that $\rho[\cdot]$ is concave, it holds that $\rho_{\ut}[\V_0^{{\vpi}^*}] \geq \rho[\V_0^{{\vpi}^*}]$ since for all $\vv\in\mcV$, we have that 
\[\rho[\vv] = \frac{1}{N!}\sum_{Q\in\G} \rho[Q\vv] \leq \rho[\frac{1}{N!}\sum_{Q\in\G} Q\vv] = \rho[\frac{1}{N}\sum_{n=1}^N v_n \vone] = \frac{1}{N}\sum_{n=1}^N v_n = \rho_{\ut}[\vv],\]
where we use permutation invariance of $\rho$, followed with its concavity and its contant vector invariance.
The inequalities in (\ref{ineq:ggf-extension}) should therefore all reach equality:
\begin{align*}
\rho[\V_0^{\vpi^*}] = \rho\left[\V_0^{\bpi^*_{\ut}}\right] = \rho_{\ut}\left[\V_0^{\vpi^*_{\ut}}\right].
\end{align*}
The rest of the argument follows exactly as in the proof of Theorem \ref{thm:eqv} (see Appendix \ref{A:eqv}).\hfill $\square$

Now, we comment that the expected utility model $\rho[\vv] = u^{-1}\left({\frac{1}{N}}\sum_{n=1}^N u(v_n)\right)$, where $u(\cdot)$ is a monotone and concave function, satisfies the three properties defined in Corollary \ref{cr:utilitarian} and is a natural framework to measure fairness in resource allocation problems {as discussed in \cite{bertsimas2012efficiency}}. The concavity of $u(\cdot)$ reflects a decreasing marginal utility on the allocated resource to an individual. This property promotes equitable distributions of resources by discouraging disparities in utility. A notable instance of this model is $\alpha$-fairness \citep{mo2000fair, ju2023achieving}, which is parameterized by $\alpha > 0$ and takes the form
\[u_{\alpha}(v) :=\left\{\begin{array}{cl} \log(v) &\mbox{if $\alpha =1$,}\\\frac{v^{1-\alpha}}{1-\alpha} &\mbox{if $\alpha \neq 1$.}\end{array}\right.\]
The domain of $\rho$ is restricted to non-negative if $\alpha\neq 1$ and strictly positive otherwise.
This function covers a range of fairness objectives, from the proportional fairness ($\alpha$ = 1) to the max-min fairness ($\alpha \rightarrow \infty$).

\section{COUNT AGGREGATION MDP}\label{def:countM}
The exact form of the count aggregation MDP is obtained as follows.

\textbf{Feasible Action} \quad The set of feasible actions in state $\x$ is defined as $$\cA(\x):= \{\u \mid \sum_{s \in \S} \sum_{a \in \A} d_k(a) u_{s,a} \leq b_k, \forall k \in \gK \;; \;\sum_{a \in \A} u_{s,a} = x_s, \forall s \in \S \}.$$

\textbf{Reward Function} \quad The average reward for all {sub-MDPs} is defined as $$\bR(\x, \u) = \frac{1}{N}\sum_{s\in \S} \sum_{a\in \A} u_{s,a} \cdot r(s,a).$$

\textbf{Transition Probability} \quad The transition probability $\cP(\x' | \x, \u)$ is the probability that the number of sub-MDPs in each state passes from $\x$ to $\x'$ given the action counts $\u$. We define the pre-image $f^{-1}(\x')$ as the set containing all elements $\s^\prime \in \Sn$ that map to $\x'$, then $\cP(\x'|\x, \u) = \sum_{\s^\prime \in f^{-1}(\x')} \Pn( \s^\prime | f^{-1}(\x), g_s^{-1}(\u))$.

Given the equivalence of transitions within the pre-image set, for an arbitrary state-action pair $(\s, \a) \in \phi^{-1}(\x, \u)$, the probability of transitioning from $\x$ to $\x'$ under action $\u$ is the sum of the probabilities of all the individual transitions in the original space that correspond to this count aggregation transition. By using the transition probability in the product space, we obtain
\begin{equation}
 \cP(\x'|\x, \u)
 = \sum_{\s^\prime \in f^{-1}(\x')} \Pn( \s^\prime | \s, \a)
 = \sum_{\s^\prime \in f^{-1}(\x')} \prod \limits^{N}_{n=1} \P_n(s_n^\prime | s_n, a_n),
\label{eq:reduced-transition}
\end{equation}
for any $(\s,\a)$ such that $\x=f(\s)$ and $\u=g_s(\a)$.

\textbf{Initial distribution} \quad By using a state count representation for symmetric weakly coupled MDPs, we know that $\vone^\top \x = N$, so the cardinality of the set can be obtained through multinomial expansion that $(s_1 + s_2 + \cdots + s_N)^{S} = \sum_{x_1 + x_2 + \cdots + x_{S} = N} \frac{N!}{x_1!x_2!\cdots x_{S}!} s_1^{x_1}s_2^{x_2}\cdots s_N^{x_{S}}$. {Intuitively, the term $s_1^{x_1}s_2^{x_2}\cdots s_N^{x_{S}}$ can represent distributing $N$ identical objects (in this case, sub-MDPs) into $S$ distinct categories (corresponding to different states). Thus, for each state count $\x$, the number of distinct ways to distribute $N$ sub-MDPs into $S$ states such that the counts match $\x$ is given by the multinomial coefficient} $|f^{-1}(\x)| = \frac{N!}{x_1!x_2!\cdots x_{S}!}$. Given the initial distribution $\vmu$ is permutation invariant, the probability of starting from state $\x$ in the initial distribution is
\begin{equation}
    \cmu(\x) = \sum_{\s \in f^{-1}(\x)} \vmu(\s)
     = |f^{-1}(\x)| \cdot \vmu(\bar{\s}) = \frac{N!}{x_1!x_2!\cdots x_{S}!} \cdot \vmu(\bar{\s}), \forall \x,
\label{eq:reduced-mu}
\end{equation}
for any $\bar{\s}$ such that $f(\bar{\s})=\x$.

\section{EXACT APPROACHES BASED ON LINEAR PROGRAMMING} 
\subsection{Optimal Solutions to the GGF-WCMDP Problem} \label{A:dual-lp}

First, we recall the dual linear programming (LP) methods to solve the MDP with discounted rewards when the transition and reward functions are known. The formulation is based on the Bellman equation for optimal policy, and is derived in Section 6.9.1 in detail by \cite{puterman2014markov}.

The dual LP formulation for addressing the multi-objective joint MDP can be naturally extended to the context of vector optimization:
\begin{equation}
    \begin{array}{rl}
         \operatorname{v-max} & \sum \limits_{\s \in \Sn} \sum \limits_{\a \in \An} \r(\s, \a) q(\s, \a) \\
         \mathrm{s.t.} & \sum\limits_{\a \in \An} q(\s, \a) -  \gamma \sum \limits_{\s^\prime \in \Sn} \sum \limits_{\a \in \An} 
         q(\s^\prime, \a) \Pn(\s | \s^\prime, \a)  =  \vmu(\s), \, \forall \s \in \Sn \\
         & q(\s, \a) \geq 0 \hfill \forall \s \in \Sn,  \forall \a \in \An
    \end{array},
    \label{vecdual}
\end{equation}
where any $\vmu(\s) > 0$ can be chosen, but we normalize the weights such that $\sum \limits_{\s \in \Sn} \vmu(\s) = 1$ can be interpreted as the probability of starting in a given state $\s$. 

We can now formally formulate the fair optimization problem by combining the GGF operator (Section \ref{sec:GGF}) and the scalarizing function on the reward vector in (\ref{vecdual}):

\begin{equation}
    \begin{array}{rl}
         \max & \GGF_{\vw}[\vv]\\
         \mathrm{s.t.} & \vv = \sum \limits_{\s \in \Sn} \sum \limits_{\a \in \An} \r(\s, \a) q(\s, \a) \\
         & \sum\limits_{\a \in \An} q(\s, \a) -  \gamma \sum \limits_{\s^\prime \in \Sn} \sum \limits_{\a \in \An} 
         q(\s^\prime, \a) \Pn(\s | \s^\prime, \a)  =  \vmu(\s), \, \forall \s \in \Sn \\
         & q(\s, \a) \geq 0 \hfill \forall \s \in \Sn,  \forall \a \in \An
    \end{array}.
    \label{ggf-mdp}
\end{equation}
By adding a permutation matrix $\gQ$ to replace the permutation applied to the index set, $\GGF_{\vw}[\vv]$ is equivalently represented as
\begin{equation}
\GGF_{\vw}[\vv]=\inf_{\bm{\gQ}:\bm{\gQ}\geq 0,\sum_i \gQ_{ij} = 1, \forall j,\;\sum_j \gQ_{ij} = 1, \forall i} \sum_{ij} w_i \gQ_{ij} v_j.
\label{ggf-s}
\end{equation}
This reformulation relies on $w_1\geq w_2\geq \dots \geq w_N$ to confirm that at the infimum we have $\min\limits_\sigma \sum \limits_{n=1}^N {w}_n {v}_{\sigma(n)}$. Indeed, if $w_1$ is not assigned to the lowest element of $\vv$, then one can get a lower value by transferring the assignment mass from where it is assigned to that element to improve the solution. This form is obtained through LP duality on (\ref{ggf-s}):
\[\sup_{\boldsymbol{\nu},\boldsymbol{\lambda}:\lambda_i + \nu_j \leq w_i v_j,\forall i,j} \sum_{i=1}^N \lambda_i + \sum_{j=1}^N \nu_j,\]
which leads to
\[
    \begin{array}{rll}
         \max \limits_{\boldsymbol{\nu},\boldsymbol{\lambda}, \boldsymbol{q}} & \sum \limits_{i=1}^N \lambda_i + \sum \limits_{j=1}^N \nu_j\\
         \mathrm{s.t.} & \lambda_i + \nu_j \leq w_i v_j & \forall i,j=1,\dots,N\\
    \end{array}.
\]

Dual variable vectors are denoted by $\bm{\lambda}$ and $\bm{\nu}$. Combining the constraints in (\ref{ggf-mdp}), we can get the complete dual LP model with the GGF objective (GGF-LP) in (\ref{ggf-mdp-d}).

\subsection{Solving Count Aggregation MDP by the Dual LP Model} \label{A:count-dual-lp}

Since the exact model for the count aggregation MDP $\cM$ is obtained (Appendix \ref{def:countM}), a dual LP model is formulated following Section 6.9.1 of \cite{puterman2014markov}, but with count aggregation representation to solve (\ref{eq:ggf-eq}):
\begin{equation}
    \begin{array}{rl}
         \max &  \sum \limits_{\x \in \cS} \sum \limits_{\u \in \cA} \bR(\x, \u) q_{\phi}(\x, \u) \\
         \mathrm{s.t.} & \sum\limits_{\u \in \cA} q_{\phi}(\x, \u) -  \gamma \sum \limits_{\x' \in \cS} \sum \limits_{\u \in \cA} 
         q_{\phi}(\x', \u) \Pn_{\phi}(\x | \x', \u)  =  \cmu(\x), \, \forall \x \in \cS \\
         & q_{\phi}(\x, \u) \geq 0 \hfill \forall \x \in \cS,  \forall \u \in \cA
    \end{array}.
    \label{eq:count-dlp}
\end{equation}
By choosing the initial distribution as $\cmu$, the optimal solution $q_{\phi}(\x, \u), \forall \x,  \u$ is equivalent to the optimal solution to the corresponding weakly coupled MDP under the transformation $\phi$.

\section{MODEL SIMULATOR} \label{apdx:algorithm}
{In the learning setting, the deep RL agent interacts with a simulated environment through a model simulator (Algorithm \ref{alg:simulation}), which leads to the next state $\x'$ and the average reward $\bR$ across all coupled MDPs.}

\begin{algorithm}[htb]
\caption{Simulation of Transition Dynamics}
\label{alg:simulation}
\begin{algorithmic}
\STATE {\bfseries Input:} count state $\x$, count action $\u$, transition probability $p(s' | s, a)$ and reward function $r(s, a)$.
\STATE \textbf{Initialize}: next state $\x' \leftarrow \bm{0}$, average reward $\bR \leftarrow 0$
\FOR{$s = 1, \dots, S$}
\FOR{$a = 1, \dots, A$}
\WHILE{$u_{s, a} > 0$}
\STATE Sample the next state index $s' \in [S]$ according to the probability distribution $p(\cdot | s, a)$
\STATE $x'_{s'} \leftarrow x'_{s'} + 1$
\STATE $\bR \leftarrow \bR + \frac{1}{N} \cdot r(s, a)$
\STATE $u_{s, a} \leftarrow u_{s, a} - 1$
\ENDWHILE
\ENDFOR
\ENDFOR
\STATE {\bf Return:} next state $\x'$, average reward $\bR$
\end{algorithmic}
\end{algorithm}

\section{EXPERIMENTAL DESIGN} 

\subsection{Parameter Setting}\label{appendix:mrp-parameter}
This section details the construction of the components used to generate the test instances based on \cite{akbarzadeh2019restless}, including the cost function, the transition matrix, and the reset probability. This experiment uses a synthetic data generator implemented on our own that considers a system with $S$ states and binary actions ($A=2$), where the two possible actions are to operate or to replace. After generating the cost matrix of the size $\Sn \cdot \An \cdot N$, we normalize the costs to the range [0, 1] by dividing each entry by the maximum cost over all state action pairs. This ensures that the discounted return always falls within the range $[0, \frac{1}{1-\gamma}]$.

\textbf{Cost function \quad } 
The cost function $c(s)$ for $s \in [S]$ can be defined in five ways: 1) \textit{Linear}: $c(s)=s-1$, where the cost increases linearly with the state index; 2) \textit{Quadratic}: $c(s) = (s - 1)^2$ with a more severe penalty for higher states compared to the linear case; 3) \textit{Exponential}: $c(s)=e^{s-1}$, which leads to exponentially increasing costs; 4) \textit{Replacement Cost Constant Coefficient (RCCC)}: $c(s) = 1.5 (S - 1)^2$, which is based on a constant ratio of 1.5 to the maximum quadratic cost; 5) \textit{Random}: $c(s)$ is randomly generated within the range [0,1].

\textbf{Transition function \quad } The transition matrix for the deterioration action is constructed as follows. Once the machine reaches the $S$-th state, it remains in that worst state indefinitely until being successfully reset by a replacement action. For the $s$-th state $s \in [S-1]$, the probability of remaining in the same state at the next step is given by a model parameter $p_m \in [0, 1]$, and the probability of transitioning to the $(s+1)$-th state is $1-p_m$.

\textbf{Reset probability \quad } When a replacement occurs, there is a probability $p_s$ that the machine successfully resets to the first state, and a corresponding probability $1-p_s$ of failing to be repaired and following the deterioration rule. In our experiments, we only consider a pure reset to the first state with probability 1. 

\subsection{Chosen Cost Structures}\label{appendix:cost}
We consider two cost models to reflect real-world maintenance and operation dynamics:

\begin{enumerate}[label=(\roman*), leftmargin=*]
    \item \textit{Exponential-RCCC}: In this scenario, operational costs increase exponentially with age, and exceed replacement costs in the worst state to encourage replacements. This scenario fits the operational dynamics of transportation fleets, such as drone batteries, where operational inefficiencies grow rapidly and can lead to significant damage to the drones.
    \item \textit{Quadratic-RCCC}: In contrast to \textit{scenario ii)}, operational costs increase quadratically with machine age, while replacement costs remain constant and always higher than operational costs. This setup is typical for high-valued machinery, where replacement costs can be significant compared to operational expenses. 
\end{enumerate}

{
\subsection{Comparison of CP-DRL Algorithms}\label{appendix:CR-DRL}
As presented in Tables \ref{tab:cp-drl1} and \ref{tab:cp-drl2}, the PPO algorithm consistently shows high-quality performance with low variance compared to TD3 and SAC. This motivated us to choose PPO to implement the CP-DRL algorithm.}

\begin{table}[htb]
\centering
\caption{GGF Scores (Exponential-RCCC)}
\begin{tabular}{ccccc}
\hline
&$N$=2	& $N$=3	&$N$=4	&$N$=5\\
\hline
PPO	&\textbf{14.11} $\pm$ 0.01	&\textbf{13.89} $\pm$ 0.14	&\textbf{13.59} $\pm$ 0.10	&13.28 $\pm$ 0.03 \\
TD3	&11.83 $\pm$ 2.74&	12.30 $\pm$ 1.14&	12.63 $\pm$ 0.23&	12.62 $\pm$ 0.12\\
SAC	&14.00 $\pm$ 0.05	&13.76 $\pm$ 0.03	&13.50 $\pm$ 0.05	&\textbf{13.30} $\pm$ 0.02\\
\hline
\end{tabular}
\label{tab:cp-drl1}
\end{table}

\begin{table}[htb]
\centering
\caption{GGF Scores (Quadratic-RCCC)}
\begin{tabular}{ccccc}
\hline
&$N$=2	& $N$=3	&$N$=4	&$N$=5\\
\hline
PPO	&\textbf{16.14} $\pm$ 0.00	&\textbf{16.05} $\pm$ 0.00	&\textbf{15.94} $\pm$ 0.00	&\textbf{15.86} $\pm$ 0.02\\
TD3	&15.98 $\pm$ 0.03	&15.75 $\pm$ 0.12	&15.56 $\pm$ 0.24	&15.51 $\pm$ 0.34\\
SAC	&16.06 $\pm$ 0.01	&15.75 $\pm$ 0.08	&15.69 $\pm$ 0.04	&15.28 $\pm$ 0.05\\
\hline
\end{tabular}
\label{tab:cp-drl2}
\end{table}

\subsection{Hyperparameters} \label{appendix:hyperparameter}
In our experimental setup, we chose PPO algorithm to implement the count-proportion based architecture. The hidden layers are fully connected and the Tanh activation function is used. There are two layers, with each layer consisting of 64 units. The learning rate for the actor is set to $5 \times 10^{-4}$ and the critic is set to $3 \times 10^{-4}$. {The Vanilla-DRL baseline uses the same network architecture as PPO, with two hidden layers of 64 neurons each, but with a softmax output layer for its stochastic policy.}

\subsection{Additional Results on LP Solving Times} \label{appendix:solving-time}
The results in Tables \ref{tab:statistics-GGF-LP} and \ref{tab:statistics-cdlp} provide details on solving the GGF-LP model and the count dual LP model on the \textit{Quadratic-RCCC} instances as the number of machines $N$ increases from 2 to 7. The state size is set to $S=3$, the action size to $A=2$, and the resource to $b=1$. The first block of the tables shows the number of constraints and variables. The second block provides the model solving times, with the standard deviations listed in parentheses. The results are based on 5 runs. Notice that the LP solve time inludes pre-solve, the wallclock time, and post-solve times. The wallclock time is listed separately to highlight the difference from the pure LP Solve time, but not included in the total time calculation.

\begin{table}[htb]\renewcommand{\arraystretch}{1.2}
\centering
\caption{Statistics for GGF-LP Model (\ref{ggf-mdp-d})}
\label{tab:statistics-GGF-LP}
\begin{tabular}{>{\centering\arraybackslash}p{0.18\linewidth}>{\centering\arraybackslash}p{0.1\linewidth}>{\centering\arraybackslash}p{0.1\linewidth}>{\centering\arraybackslash}p{0.1\linewidth}>{\centering\arraybackslash}p{0.1\linewidth}>{\centering\arraybackslash}p{0.1\linewidth}>{\centering\arraybackslash}p{0.1\linewidth}}
\hline
    & {$N$ = 2} & {$N$ = 3} & {$N$ = 4} & {$N$ = 5} & {$N$ = 6} & $N$ = 7 \\ \hline
\# Constraints      & 13& 36& 97& 268& 765&         2236\\
\# Variables        & 31& 114& 413& 1468& 5115&         17510\\ \hline
Data Build (s)      & 0.0019 (0.00)& 0.0085 (0.00)& 0.1076 (0.01)& 1.3698 (0.05)&                             17.7180 (0.52)&         320.0141 (16.85)\\
LP Build (s)        & 0.0028 (0.00)& 0.0185 (0.01)& 0.1449 (0.01)& 1.4673 (0.08)& 20.7464 (4.00)&         392.0150 (33.97)\\
LP Solve (s)        & 0.0187 (0.02)& 0.0212 (0.00)& 0.1846 (0.06)& 1.2914 (0.09)& 13.0377 (0.63)&         138.1272 (2.33)\\ 
Wallclock Solve$^*$ (s) & 0.0026 (0.00)& 0.0018 (0.00)& 0.0115 (0.00)& 0.0493 (0.00)& 0.7849 (0.16)&         13.3167 (0.19)\\
LP Extract (s)      & 0.0022 (0.00)& 0.0019 (0.00)& 0.0044 (0.00)& 0.0158 (0.00)& 0.0711 (0.01)&         0.2828 (0.03)\\ \hline
\textbf{Total Time (s)}               & 0.0256 (0.03)& 0.0500 (0.01)& 0.4416 (0.06)& 4.1443 (0.07)& 51.5732 (3.87)&         864.4391 (56.75)\\ \hline
\end{tabular}
\vspace{3pt}

{\raggedright \, $^*$Wall clock solve time is included in the LP Solve time. \par}
\end{table}

\begin{table}[!htb]\renewcommand{\arraystretch}{1.2}
\centering
\caption{Statistics for Count Dual LP Model (\ref{A:count-dual-lp})}
\label{tab:statistics-cdlp}
\begin{tabular}{>{\centering\arraybackslash}p{0.18\linewidth}>{\centering\arraybackslash}p{0.1\linewidth}>{\centering\arraybackslash}p{0.1\linewidth}>{\centering\arraybackslash}p{0.1\linewidth}>{\centering\arraybackslash}p{0.1\linewidth}>{\centering\arraybackslash}p{0.1\linewidth}>{\centering\arraybackslash}p{0.1\linewidth}}
\hline
    & {$N$ = 2} & {$N$ = 3} & {$N$ = 4} & {$N$ = 5} & {$N$ = 6} & $N$ = 7 \\ \hline
\# Constraints      & 6& 10& 15& 21& 28&         36\\
\# Variables        & 24& 40& 60& 84& 112&         144\\ \hline
Data Build (s)      & {0.0035} (0.00)& 0.0134 (0.00)& 0.1306 (0.01)& 1.4401 (0.01)&                             17.4080 (0.40)&         204.4550 (4.49)\\
LP Build (s)        & {0.0018} (0.00)& 0.0031
 (0.00)& 0.0056 (0.01)& 0.0081 (0.00)& 0.0297 (0.01)&         0.0501 (0.01)\\
LP Solve (s)        & {0.0375} (0.02)& 0.0362
 (0.02)& 0.0466 (0.00)& 0.0386 (0.01)& 0.0745 (0.01)&         0.1634 (0.06)\\ 
Wallclock Solve$^*$ (s) & 0.0034 (0.00)& 0.0053
 (0.00)& 0.0031 (0.00)& 0.0026 (0.00)& 0.0034 (0.00)&         0.0105 (0.00)\\
LP Extract (s)      & 0.0053 (0.00)& 0.0025
(0.00)& 0.0687 (0.00)& 0.0026 (0.00)& 0.0051 (0.00)&         0.0209 (0.00)\\ \hline
\textbf{Total Time (s)}               & 0.0481 (0.02)& 0.0552  (0.02)& 0.2515 (0.13)& 1.4894 (0.01)& 17.5173 (0.42)&         204.6893 (4.44)\\ \hline
\end{tabular}
\vspace{3pt}

{\raggedright \, $^*$Wall clock solve time is included in the LP Solve time. \par}
\end{table}

\end{document}

%% file: math_commands.tex
\usepackage{amsmath,amsfonts,bm}
\usepackage{amstext}
\usepackage{amssymb}
\usepackage{bm}
\usepackage{enumitem}

\def\eqref#1{equation~\ref{#1}}

\def\1{\bm{1}}

\def\vone{{\bm{1}}}
\def\voneN{{\bm{1}}/N}

\def\vb{{\bm{b}}}
\def\dvb{\bar{\bm{b}}}
\def\pb{\tilde{\bm{p}}}

\def\vd{{\bm{d}}}

\def\vv{{\bm{v}}}
\def\vw{{\bm{w}}}

\def\mU{{\bm{U}}}

\DeclareMathAlphabet{\mathsfit}{\encodingdefault}{\sfdefault}{m}{sl}
\SetMathAlphabet{\mathsfit}{bold}{\encodingdefault}{\sfdefault}{bx}{n}

\def\gF{{\mathcal{F}}}

\def\gK{{\mathcal{K}}}

\def\gN{{\mathcal{N}}}

\def\gQ{{\mathcal{Q}}}

\def\gT{{\mathcal{T}}}

\def\gX{{\mathcal{X}}}

\def\sI{{\mathbb{I}}}

\def\sN{{\mathbb{N}}}

\def\sR{{\mathbb{R}}}
\def\sS{{\mathbb{S}}}

\newcommand{\E}{\mathbb{E}}

\newcommand{\R}{\mathbb{R}}

\DeclareMathOperator{\GGF}{GGF}

\newcommand{\Prob}{\mathbb{P}}
\newcommand{\M}{\mathcal{M}}
\renewcommand{\S}{\mathcal{S}}
\newcommand{\A}{\mathcal{A}}
\renewcommand{\P}{p}
\renewcommand{\R}{r}

\newcommand{\Mn}{\M^{(N)}}
\newcommand{\Sn}{\S^{(N)}}
\newcommand{\An}{\A^{(N)}}

\newcommand{\Pn}{\P^{(N)}}
\newcommand{\Rn}{\bm{r}}
\newcommand\vmu{\mu^{(N)}}
\newcommand{\cM}{\M_\phi}
\newcommand{\cS}{\Sn_f}
\newcommand{\cA}{\An_{g_s}}
\newcommand{\cP}{\Pn_\phi}

\newcommand{\bR}{\bar{r}_{\phi}}
\newcommand{\cmu}{\vmu_f}
\newcommand{\ut}{U}
\newcommand{\s}{\bm{s}}
\renewcommand{\a}{\bm{a}}

\renewcommand{\r}{\bm{r}}
\newcommand{\x}{\bm{x}}
\renewcommand{\u}{\bm{u}}
\newcommand{\so}{\bar{\s}_0} 

\renewcommand{\E}{\mathbb{E}}
\newcommand{\V}{\bm{V}}
\newcommand{\bV}{\bar{V}_0}
\newcommand{\Vo}{\bm{V}_0}
\newcommand{\Vmu}{\bm{V}^{\vpi}_0}
\newcommand{\Vm}{\bm{V}^{\vpi}_0} 

\newcommand{\vpi}{\pi} 
\newcommand{\bpi}{\bar{\vpi}} 
\newcommand{\cpi}{{\vpi}_{\phi}} 
\newcommand{\ppi}{\vpi^Q} 

\newcommand{\G}{\mathcal{G}^{N}}

\newcommand{\dx}{\bar{\bm{x}}}

\newtheorem{theorem}{Theorem}[section]
\newtheorem{definition}[theorem]{Definition}
\newtheorem{corollary}{Corollary}[theorem]

\newtheorem{lemma}[theorem]{Lemma}